\documentclass{article}

\usepackage{amsmath,amsthm,amssymb,amsfonts}
\usepackage{graphicx}
\usepackage{arxiv}

\usepackage[utf8]{inputenc} 
\usepackage[T1]{fontenc}    
\usepackage{hyperref}       
\usepackage{url}            
\usepackage{booktabs}       
\usepackage{amsfonts}       
\usepackage{nicefrac}       
\usepackage{microtype}      
\usepackage{lipsum}

\usepackage{xcolor}         
\usepackage{multirow}
\usepackage{caption}
\usepackage{subcaption}
\usepackage{graphicx}
\usepackage{amsfonts, amsmath, amssymb} 
\usepackage{algorithm}
\usepackage{algorithmic}

\usepackage{wrapfig}
\usepackage{tikz}
\usepackage{array}

\usepackage{amsfonts,amssymb,amsthm,amsmath}
\usepackage{subcaption}
\usepackage{booktabs}

\usepackage{times}
\usepackage{helvet}
\usepackage{courier}
\usepackage{xcolor}

\newcommand{\modelname}{\textbf{ADATG}}
\newcommand{\base}{\textbf{SFM}} 
\newcommand{\hevit}{\textbf{HE-ViT}} 
\newcommand{\twogrid}{\textbf{Fixed-TG}}
\newcommand{\rantwogrid}{\textbf{Ran-TG}}
\newcommand{\modelone}{\textbf{ADATG-HH}}
\newcommand{\modeltwo}{\textbf{ADATG-NH}}

\newtheorem{definition}{Definition}

\title{Synergizing Multigrid Algorithms with Vision Transformer: A Novel Approach to Enhance the Seismic Foundation Model}

\author{
  Huiwen Wu~\thanks{This work has been accepted for publication in the Proceedings of the 40th AAAI Conference on Artificial Intelligence (AAAI 2026).} ~ \thanks{corresponding author}  \\
  Research Center for Scientific Data Hub \\
  Zhejiang Laboratory\\
  Hangzhou, Zhejiang, China \\
  \texttt{huiwen0820@outlook.com} \\
   \And
    Shuo Zhang~ \thanks{corresponding author} \\
    State Key Laboratory of Mathematical Sciences (SKLMS) \\
    State Key Laboratory of Scientific and Engineering Computing (LSEC) \\
    Institute of Computational Mathematics and Scientific/Engineering Computing \\
    Academy of Mathematics and Systems Science, Chinese Academy of Sciences, Beijing, China. \\
    School of Mathematical Sciences, University of Chinese Academy of Sciences, Beijing, China.  \\
    \texttt{szhang@lsec.cc.ac.cn} \\
   \And
    Yi Liu \\
    Research Center for Scientific Data Hub \\
    Zhejiang Laboratory\\
    Hangzhou, Zhejiang, China \\
    \texttt{liuyi4@zhejianglab.org} \\
   \And
    Hongbin Ye \\
    Research Center for Scientific Data Hub \\
    Zhejiang Laboratory\\
    Hangzhou, Zhejiang, China \\
    \texttt{zjuhongbinye@gmail.com} \\
}

\begin{document}
\maketitle

\begin{abstract}
Due to the emergency and homogenization of Artificial Intelligence (AI) technology development, transformer-based foundation models have revolutionized scientific applications, such as drug discovery, materials research, and astronomy. However, seismic data presents unique characteristics that require specialized processing techniques for pretraining foundation models in seismic contexts with high- and low-frequency features playing crucial roles. Existing vision transformers (ViTs) with sequential tokenization ignore the intrinsic pattern and fail to grasp both the high- and low-frequency seismic information efficiently and effectively.
This work introduces a novel adaptive two-grid foundation model training strategy (\modelname) with Hilbert encoding specifically tailored for seismogram data, leveraging the hierarchical structures inherent in seismic data. Specifically, our approach employs spectrum decomposition to separate high- and low-frequency components and utilizes hierarchical Hilbert encoding to represent the data effectively.
Moreover, observing the frequency principle observed in ViTs, we propose an adaptive training strategy that initially emphasizes coarse-level information and then progressively refines the model's focus on fine-level features. 
Our extensive experiments demonstrate the effectiveness and efficiency of our training methods. This research highlights the importance of data encoding and training strategies informed by the distinct characteristics of high- and low-frequency features in seismic images, ultimately contributing to the enhancement of visual seismic foundation models pretraining.
\end{abstract}

\begin{figure}
\centering 
\includegraphics[width=0.23\textwidth]{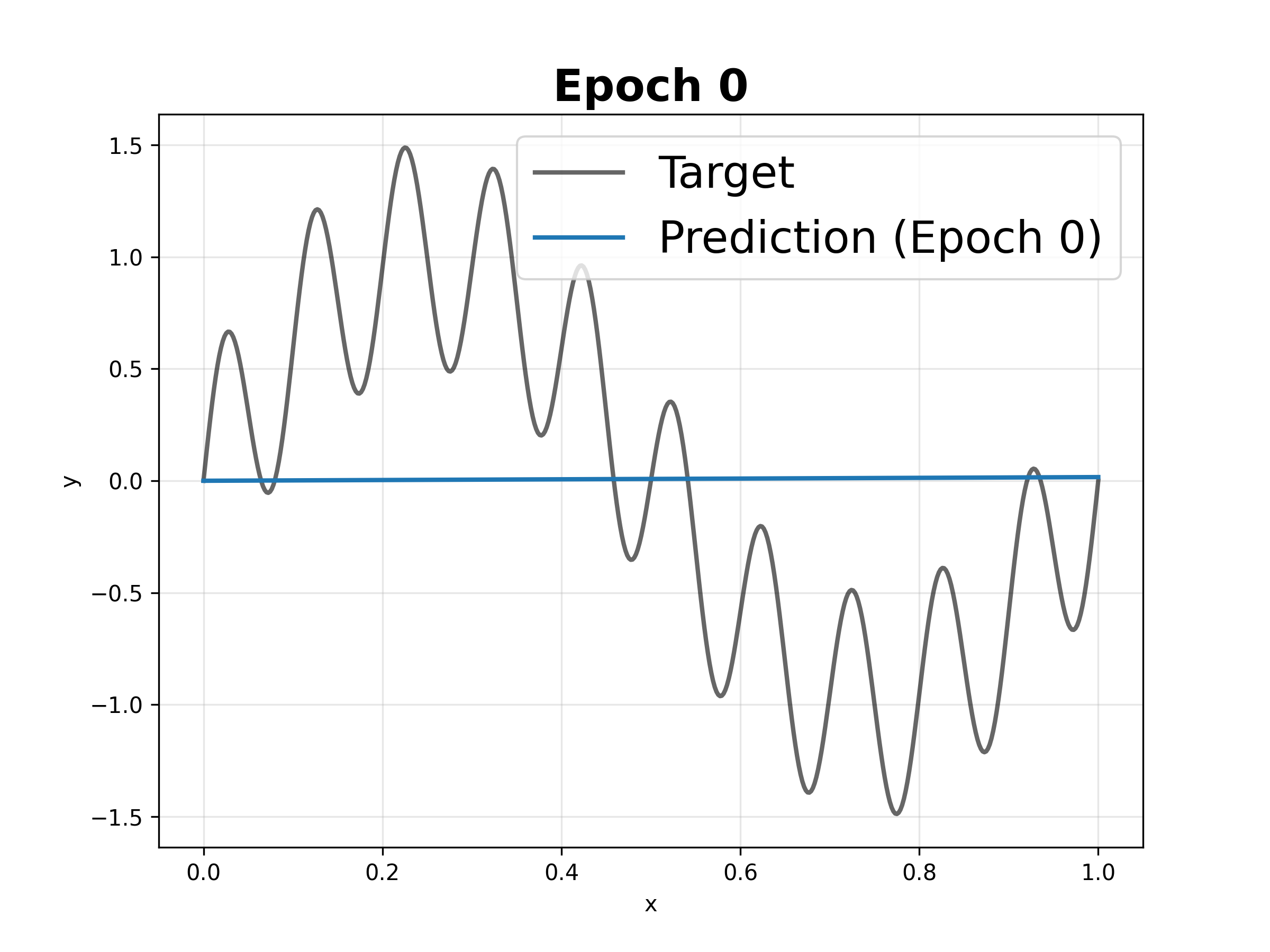}
\includegraphics[width=0.23\textwidth]{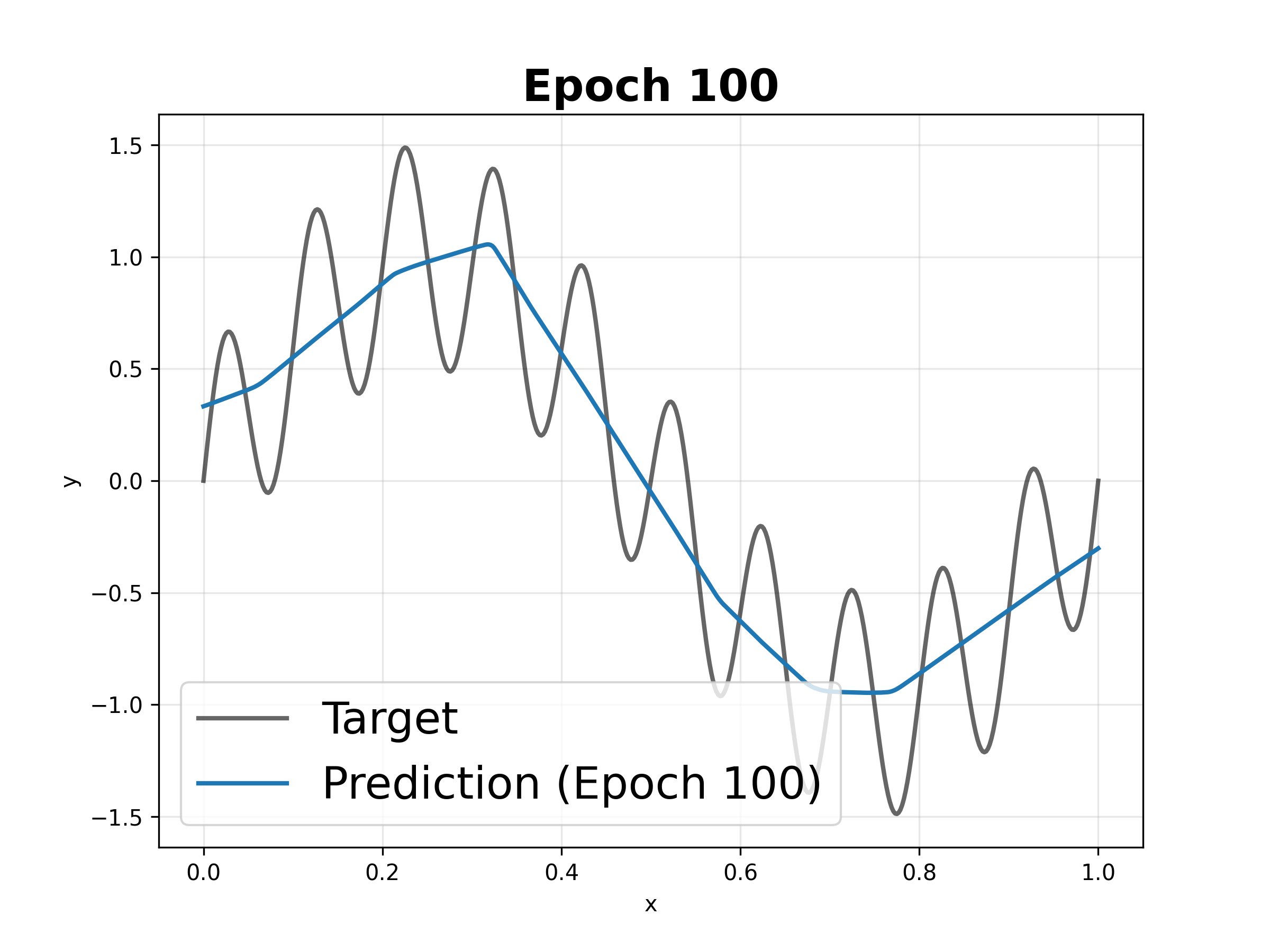}
\includegraphics[width=0.23\textwidth]{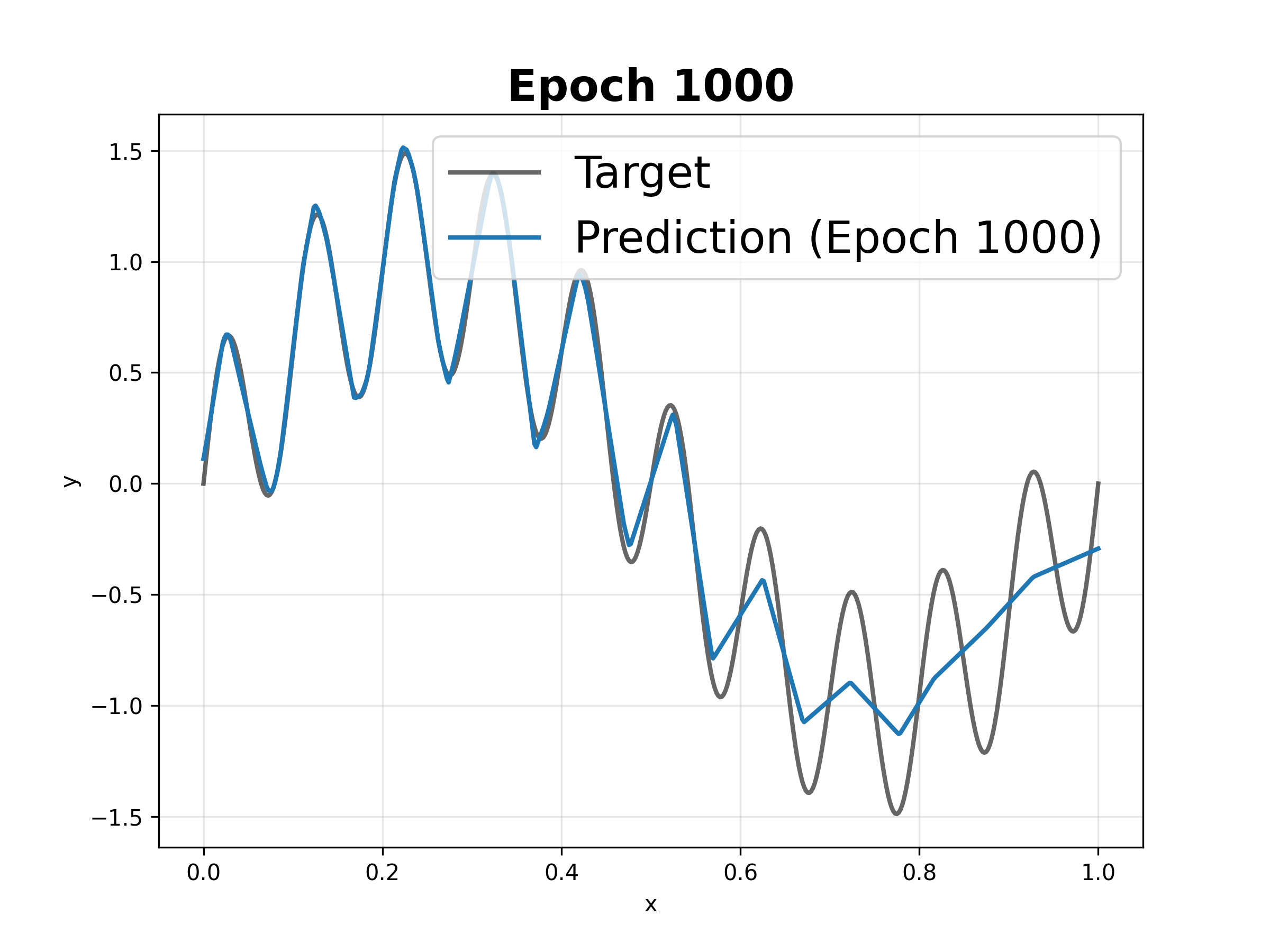}
\includegraphics[width=0.23\textwidth]{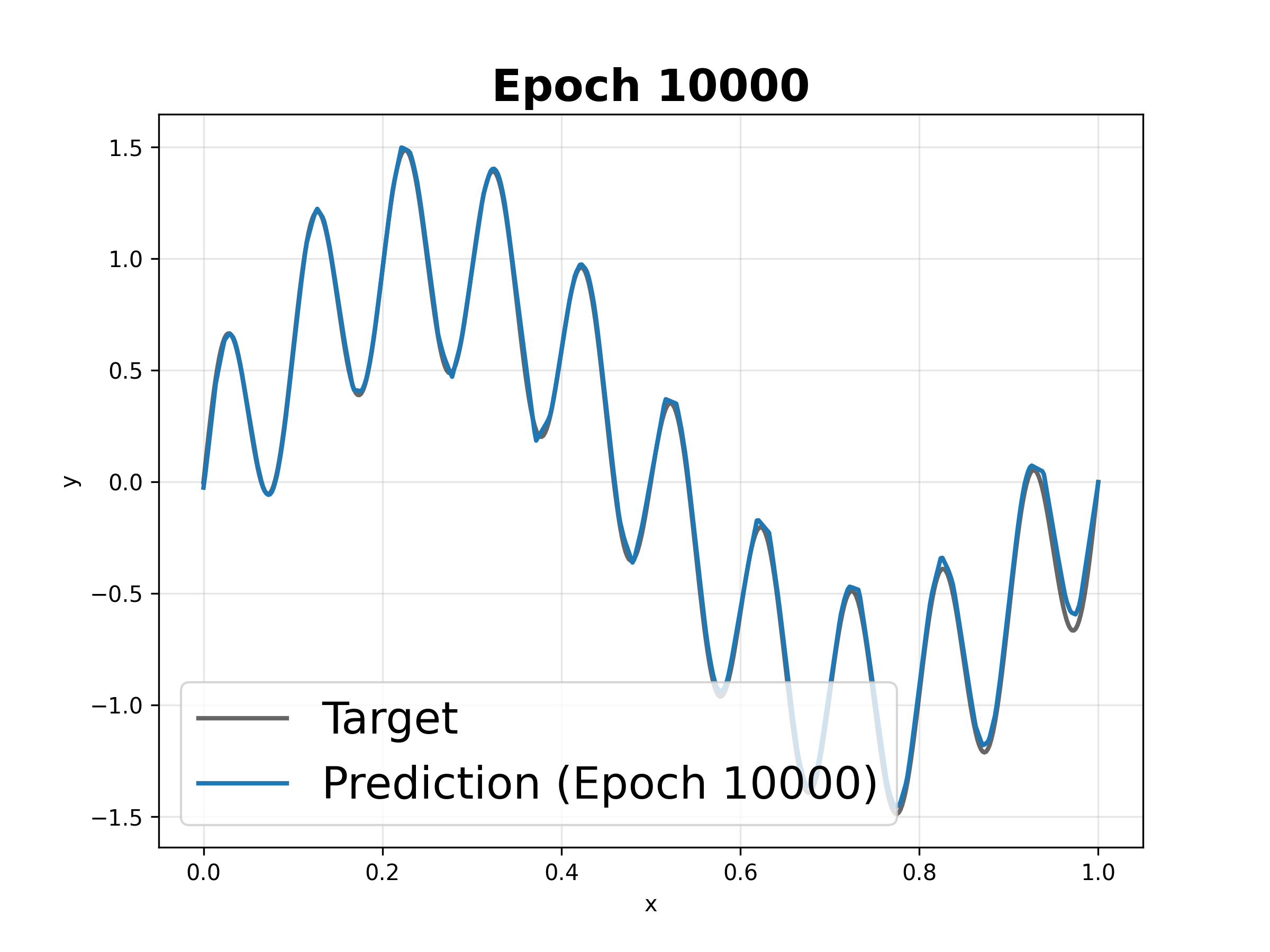}
\caption{A toy example demonstrating frequency principle.}
\label{fig:freq}
\end{figure}

\begin{figure*}[t!]
 \centering
 \includegraphics[width=\textwidth]{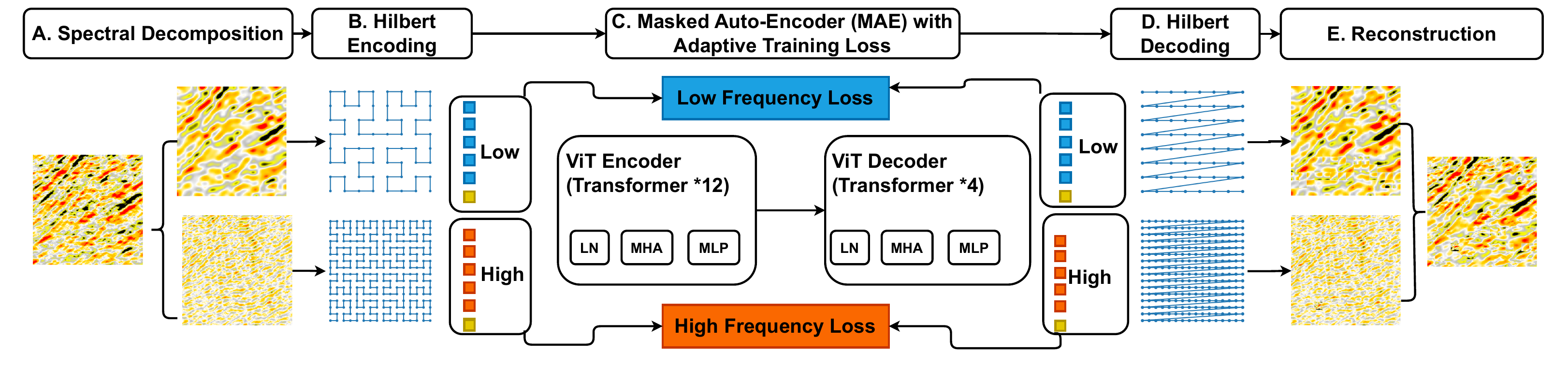}
\caption{Pipeline of Seismic Foundation Model Pretraining with \modelname.
(A) Input seismogram spectral decomposition via Fourier transform,
(B) Frequency-adaptive Hilbert encoding (fine/coarse grids for high/low frequencies),
(C) Vision Transformer training with adaptive MAE strategy,
(D) Frequency-adaptive Hilbert decoding with inverse transformer matrix, 
(E) Merged reconstruction of difference frequency components. }
\label{fig:overall}
\end{figure*}

\section{Introduction}

The rapid development of transformer-based models has enabled a wide range of AI-driven applications in complex scientific tasks. This includes ligand coupling for drug discovery~\cite{zhang2023efficient}, materials discovery~\cite{pyzer2025foundation}, and mathematical proof~\cite{xin2024deepseek}. Recently, significant attention has been paid to training seismic foundation models~\cite{sheng2025seismic, si2024seisclip, liu2024seislm, li2024seist} using large-scale, globally sourced seismic data. 
These pre-trained seismic models can be effectively utilized in various downstream tasks, including seismic facies classification, geobody segmentation, seismic image denoising, and full seismic inversion~\cite{sheng2025seismic}. The development and training of seismic foundation models are essential for advancing both scientific research and practical engineering applications in geophysics.

Due to the intrinsic characteristics of the seismic data, the seismogram can be decomposed into high-frequency and low-frequency components. The high-frequency components are typically associated with rapid changes in ground motion and shorter wavelengths~\cite{yilmaz2001,sheriff1995}. In contrast, low-frequency components correspond to slower changes in ground motion and are linked to longer wavelengths~\cite{pratt1999,virieux2009}.  
These high-frequency and low-frequency components play distinct but complementary roles in seismic analysis. From the perspective of waveform complexity, high frequencies contribute to the sharpness of seismic phases, such as the arrivals of P- and S-waves. In contrast, low frequencies predominantly shape the overall envelope of signals, particularly surface waves~\cite{aki2002quantitative,stein2003introduction}. In event discrimination tasks, the high-frequency to low-frequency energy ratio can help distinguish between tectonic earthquakes and explosions, as explosions tend to generate more high-frequency energy~\cite{fichtner2010,tarantola2005}.

\begin{quote}
\textit{Thus, it is crucial to develop a unified foundation model effectively capturing both high-frequency and low-frequency components simultaneously.}
\end{quote}

Training foundation models to effectively leverage high- and low-frequency features presents several challenges. First, in the traditional Vision Transformer (ViT) model, images are divided into $16 \times 16$ tokens, which are processed sequentially (Figure~\ref{fig:hilbert_encoding} (a)). This sequential encoding can disrupt intrinsic patterns within the seismogram.
Second, because the seismogram contains both high- and low-frequency information, both of which are crucial for downstream tasks, existing ViT models often fail to capture the fine details represented by the high-frequency components and perform inadequately when reconstructing low-frequency components.
Lastly, motivated by the frequency principle (Figure~\ref{fig:freq_prin}) in which models preferentially learn low-frequency components before high-frequency ones during training — has been well-established in prior work~\cite{xu2019frequency}. 
Our experiments demonstrate that this principle also governs Vision Transformers (ViTs), as detailed in Sect.~\ref{sec:frequency_principle}. The existing ViT training paradigm which ignores this fact fails to acquire low- and high- frequency features accurately. 

To address the challenges outlined above, we have enhanced the training approach for seismic foundation models in several key ways.
First, we developed a Hilbert encoder for ViT to encode seismogram data according to its intrinsic patterns.
Next, we decompose the high- and low-frequency features of the input seismogram using discrete Fourier transforms and process them with different-scale Hilbert encoding. For low-frequency features, we apply the standard Hilbert encoder, while for high-frequency features, we employ a refined Hilbert encoder.
Finally, motivated by this observation, we propose a two-grid adaptive frequency decomposition strategy: the model prioritizes high-frequency features in the early training stages before shifting focus to low-frequency refinement later (Sect.~\ref{sec:ada_freq}).
The main contributions are listed below. 

\begin{itemize}
\item We propose a frequency decomposition method via a discrete Fourier transform to decompose the seismogram into high- and low-frequency components. 
\item We design a novel two-grid Hilbert encoding methods with coarse-level Hilbert encoding for the low-frequency feature and fine-level Hilbert encoding for high-frequency feature.
\item Experimental results verifies the Frequency Principle for Transformer architecture. We further design the adaptive training strategy inspired by the Frequency Principle. 
\end{itemize}

\begin{figure*}[htbp]
     \begin{subfigure}[b]{0.16\textwidth}
         \centering
         \includegraphics[width=\textwidth]{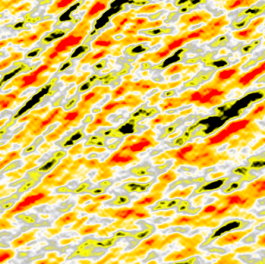}
         \caption{\textbf{Original}}
         \label{fig:original}
     \end{subfigure}
     \begin{subfigure}[b]{0.16\textwidth}
         \centering
         \includegraphics[width=\textwidth]{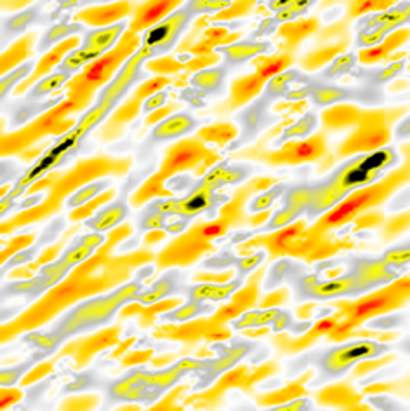}
         \caption{\base}
         \label{fig:base}
     \end{subfigure}
     \begin{subfigure}[b]{0.16\textwidth}
         \centering
         \includegraphics[width=\textwidth]{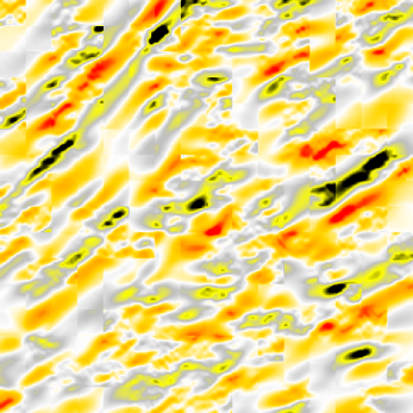}
         \caption{\hevit}
         \label{fig:hilbert}
     \end{subfigure}
     \begin{subfigure}[b]{0.16\textwidth}
         \centering
         \includegraphics[width=\textwidth]{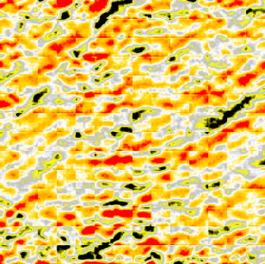}
         \caption{\rantwogrid}
         \label{fig:ran_twogrid}
     \end{subfigure}
     \begin{subfigure}[b]{0.16\textwidth}
         \centering
         \includegraphics[width=\textwidth]{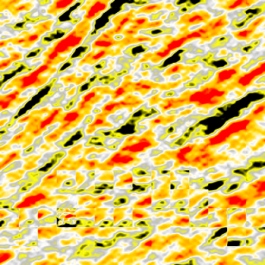}
         \caption{\modelone}
         \label{fig:ada_twogrid_chfh}
     \end{subfigure}
     \begin{subfigure}[b]{0.16\textwidth}
         \centering
         \includegraphics[width=\textwidth]{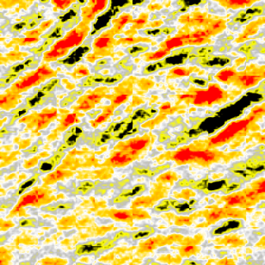}
         \caption{\modeltwo}
         \label{fig:ada_twogrid_cnfh}
     \end{subfigure}
\caption{Reconstructed images using different pretrained Seismic Foundation Models (SFM). From left to right: the original image, reconstruction using the base Vision Transformer (ViT) architecture~\cite{sheng2025seismic}, the Hilbert Encoding ViT (\hevit), the randomized Two-grid method (\rantwogrid), and two variants of \modelname: one incorporating both high and low frequency components with Hilbert Encoding (\modelone), and another using only high-frequency components (\modeltwo). All methods are applied under identical evaluation conditions to enable direct visual comparison.}
\label{fig:compare}
\end{figure*}

\section{Related Work}

\subsection{Seismic Foundation Models}
%
Recently, we encountered training foundation models for seismic data and transferred the learned knowledge to downstream tasks. 
For example, 
in SFM~\cite{sheng2025seismic}, the authors employ self-supervised learning to pretrain a Transformer-based seismic foundation model to produce all-purpose seismic features that can be applied to various downstream tasks.
In SeisCLIP~\cite{si2024seisclip}, the authors propose aa seismology foundation model trained through contrastive learning from multimodal data. 
In SeisLM~\cite{liu2024seislm}, the authors propose a foundation seismic foundation model with input data from seismic waveform signals and trained using self-supervised contrastive loss, similar to language models. 
%
%
However, since most seismic foundation models process the time-frequency seismic spectrum as input data, few works treat the seismogram as images and processing using computer vision techniques~\cite{sheng2025seismic}.
Here, we focus on the image processing approach and compare it with the SFM~\cite{sheng2025seismic}.

\subsection{Masked Image Modeling}

In \cite{dosovitskiy2020image}, the authors first apply the Transformer architecture to computer vision tasks by tokenizing input images into $16 \times 16$ patches.
The Masked Auto-Encoder (MAE) approach, introduced in~\cite{he2022masked}, involves randomly masking patches of the input image and reconstructing missing pixels. The encoder-decoder structure is asymmetric, with the encoder accessing only the unmasked patches and a lightweight decoder reconstructing the original image from the latent representation and mask tokens.
In \cite{li2022uniform}, the authors propose uniform masking, which enables MAE pre-training for pyramid-based vision transformers.
To enhance MAE efficiency, \cite{liu2023mixmae} replaces the masked tokens of one image with visible tokens of other images and trains dual reconstruction neural networks to reconstruct the two images from the mixed ones.
In this work, we combine spectrum decomposition with the MAE learning approach proposed in~\cite{he2022masked} and apply it to large-scale seismogram pre-training.

\subsection{Space-filling Curve}
A space-filling curve is a continuous function with endpoints whose domain is the unit interval $[0, 1]$. Several studies have applied space-filling curves to tasks in image processing and spatial information encoding.
In~\cite{chen2007new}, the author develops iterative Hilbert encoding and decoding methods based on the replication process of the Hilbert matrix, demonstrating reduced time and space complexity, both bounded by a constant.
In~\cite{moon2001analysis}, the authors investigate the clustering properties of the Hilbert space-filling curve by deriving closed-form formulas for the number of clusters within a given query region of arbitrary shape.
In~\cite{bhupati2019classification}, the author applies space-filling curves to the fMRI brain activation map task by fMRI for medical image processing and classification.
In this work, because of the intrinsic structure of seismogram data, we adopted the Hilbert curve to encode vision tokens.

\begin{figure*}[t!]
     \begin{subfigure}[b]{0.245\textwidth}
         \centering
         \includegraphics[width=\textwidth]{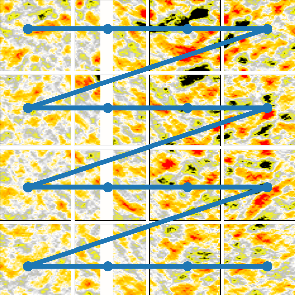}
         \caption{\textbf{Base Encoding}}
         \label{fig:base_1}
     \end{subfigure}
     \begin{subfigure}[b]{0.245\textwidth}
         \centering
         \includegraphics[width=\textwidth]{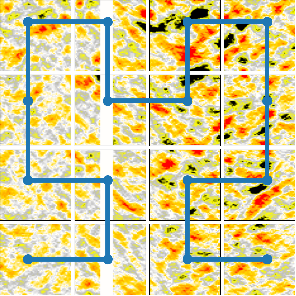}
         \caption{\textbf{Hilbert Encoding}}
         \label{fig:hilbert_1}
     \end{subfigure}
      \begin{subfigure}[b]{0.245\textwidth}
     \centering
     \includegraphics[width=\textwidth]{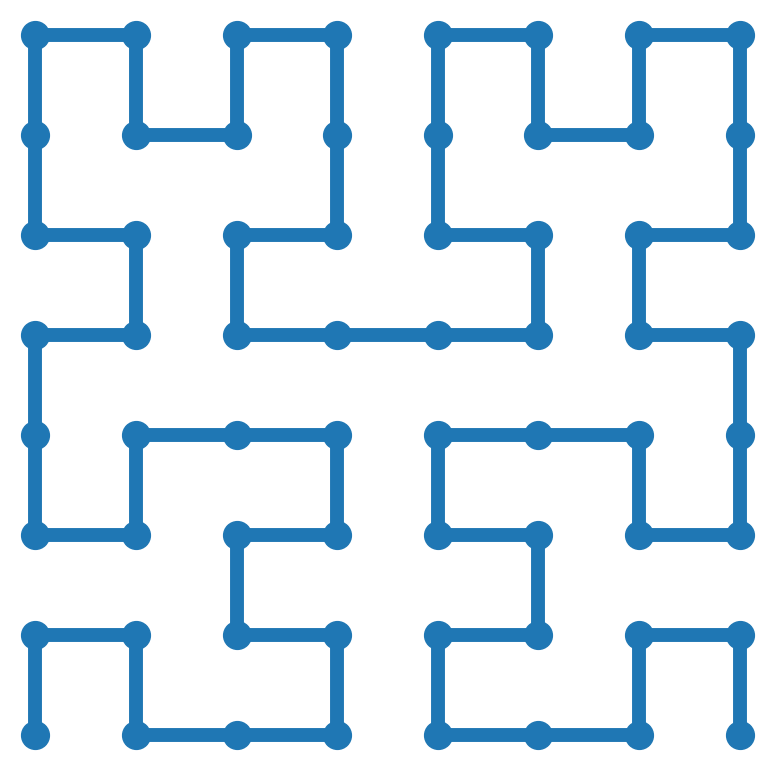}
     \caption{\textbf{Coarse Grid}}
     \label{fig:coarse}
 \end{subfigure}
 \begin{subfigure}[b]{0.245\textwidth}
     \centering
     \includegraphics[width=\textwidth]{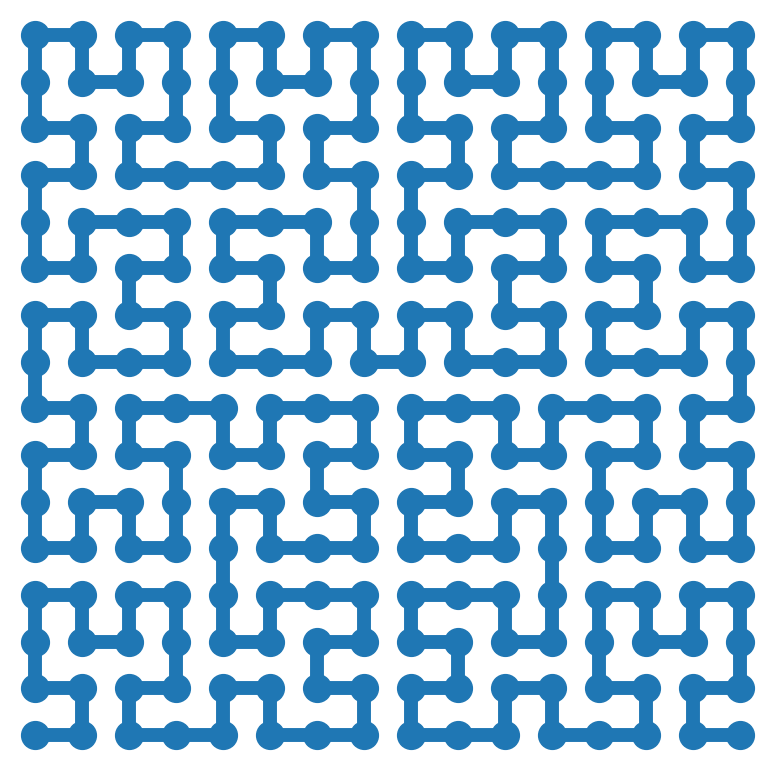}
     \caption{\textbf{Fine Grid}}
     \label{fig:fine}
 \end{subfigure}
\caption{Hilbert Encoding (Left Two) and Twogrid Hilbert Encoding (Right Two).}
\label{fig:hilbert_encoding}
\end{figure*}

\section{Methodology}


\subsection{Spectral Decomposition}
\label{sec:meth_freq}

First, we describe the spectrum decomposition via discrete Fourier transform (DFT) in matrix form. 
\begin{definition}[Discrete Fourier Transform in Matrix Form~\cite{trefethen1997numerical}]
Let $\omega = \exp(- \frac{2 \pi}{N} i)$ be such that
$\exp(- j k 2 \pi/{N} i) = \omega^{jk}$.
Then the $N \times N$ Fourier matrix is defined as 
$$
\mathcal{F}= 
\begin{bmatrix}
\omega^{0 \cdot 0} & \omega^{0 \cdot 1} &  \cdots & \omega^{0 \cdot (N-1)} \\ 
\omega^{1 \cdot 0} & \omega^{1 \cdot 1} & \cdots & \omega^{1 \cdot (N-1)} \\
\cdots & \cdots &  \cdots & \cdots \\
\omega^{(N-1) \cdot 0} & \omega^{(N-1) \cdot 1} & \cdots & \omega^{(N-1) \cdot (N-1)}
\end{bmatrix}.
$$
Then the DFT in matrix form can be defined as 
$$
\hat{\mathbf{X}} = \mathbf{DFT}(X) = \mathcal{F} \mathbf{X},
$$
and the inverse discrete Fourier transform (iDFT) in matrix form is 
$$
\mathbf{X} = \mathbf{iDFT}(\hat{\mathbf{X}}) = \mathcal{F}^{-1} \hat{\mathbf{X}}. 
$$
\end{definition}

Then we apply the DFT to the input seismogram image.
\begin{eqnarray}
\label{eq:fourier_transform}
\hat{\mathbf{X}} = \mathbf{DFT} (\mathbf{X}); \quad 
\hat{\mathbf{X}}_k = \frac{1}{n} \sum_{i=1}^n \mathbf{X}_i e^{-i 2 \pi i k /n}. 
\end{eqnarray}
Next, we decompose the high-frequency and low-frequency component in the Fourier space with a given threshold $k_0$, 
\begin{eqnarray}
\label{eq:freq_decomp}
\hat{\mathbf{X}}_{\text{low}} = \sum_{k \leq k_0} \hat{\mathbf{X}}_k; \quad
\hat{\mathbf{X}}_{\text{high}} = \sum_{k > k_0} \hat{\mathbf{X}}_k. 
\end{eqnarray}
Then we applied the inverse discrete Fourier transform (iDCT) to convert the high-frequency and low-frequency components of the seismogram into the physical domain. 
\begin{eqnarray}
\label{eq:inverse_fourier_transform}
\mathbf{X}_{\text{low}} = \mathbf{iDFT} (\hat{\mathbf{X}}_{\text{low}}); \quad
\mathbf{X}_{\text{high}} =   \mathbf{iDFT} (\hat{\mathbf{X}}_{\text{high}}). 
\end{eqnarray}


\subsection{Vision Transformer (ViT)}

In this section, we introduce the traditional ViT~\cite{dosovitskiy2020image} in the pre-training model of the seismic foundation. Let’s consider an input image with dimensions \( H \times W \times C \), where \( H \), \( W \), and \( C \) represent the height, width, and number of channels, respectively. 
In the traditional ViT framework~\cite{dosovitskiy2020image}, the first step involves reshaping the input image into a sequence of flattened two-dimensional patches, resulting in a shape of $ N \times (P^2 \times C) $. Here, $N$ is the total number of patches and $P^2$ is the area of a single patch. 
The convolutional operator used in convolutional neural networks (CNN) effectively captures both locality and translationally equivalent information~\cite{krizhevsky2012imagenet}. In contrast, in the base Vision Transformer (ViT) architecture, only the final Multi-Layer Perception (MLP) captures local and translationally equivalent information, while the self-attention layer focuses on global information~\cite{dosovitskiy2020image}. This design allows ViTs to exhibit stronger generalization capabilities, although they may capture the local texture of images less effectively.


\subsection{Hilbert Encoding}

Importantly, seismogram images present significant patterns characterized by both high- and low-frequency information. The high frequency components are typically related to shallow or local events, whereas the low frequency components may correspond to deeper events~\cite{ringler2022improved,lindsey2021fiber}. Therefore, relying solely on global or local information when analyzing seismogram images can be limiting.
To effectively capture local features in the seismogram, we introduce Hilbert encoding based on a space-filling curve.
First of all, we introduce the space filling curve.

\begin{definition}{Space Filling Curve~\cite{bially1969space}}
\label{def:space}
A space-filling curve is a continuous function $f: [0,1] \rightarrow [0,1]^n $ such that for every $y \in [0, 1]^n$, there exists a point $t \in [0,1]$ where f(t) =y. 
\end{definition}

A space-filling curve has three key properties, including continuity, surjectivity, and fractal nature. The last property enables iterative construction. 
The Hilbert curve is a special example of a space-filling curve. The iterative generation formula of the Hilbert curve is defined below. 

\begin{definition}{Hilbert Curve~\cite{butz1969convergence}}
The Hilbert curve is generated by the following formula iteratively. 
\begin{itemize}
\item{Base Case, Order 1}
The first order Hilbert curve $\mathcal{H}_1 = [(0,0), (0,1), (1,1), (1,0)].$
\item{Recursive Step, Order n}
The curve $\mathcal{H}_n$ is constructed by connecting four copies of $(n-1)-$th order Hilbert curve $\mathcal{H}_{n-1}$, each rotated and reflected to ensure continuity. 
$$
\mathcal{H}_{n+1} = [\rm{Rotate}(\mathcal{H}_n, \frac{\pi}{2}], \mathcal{H}_n , \mathcal{H}_n, \rm{Rotate}(\mathcal{H}_n, \frac{3\pi}{2})].
$$
\end{itemize}
\end{definition}
The inverse of a Hilbert curve, denoted $\mathcal{H}^{-1}$, can be defined as the operation that reverses the permutation sequence generated by the Hilbert curve $\mathcal{H}$.
Using the Hilbert encoding~\cite{chen2007new}, one can further change the sequence of the patched sequence. 
Specifically, in traditional ViT, after patchification, we group the embeddings in a right-to-left, up-to-down sequence; see Figure~\ref{fig:base_1}. 
With Hilbert encoding, the sequence of embeddings is constructed based on the Hilbert curve sequence; see Figure~\ref{fig:hilbert_1}. 
%



\subsection{Twogrid Hilbert Encoding}

%



Let $X_{\text{high}}$ and $X_{\text{low}}$ be the high- and low-frequency components obtained from the spectral decomposition defined in Eq.~\eqref{eq:inverse_fourier_transform}. 
Select the order of the Hilbert curve $n_1 ,n_2 \in \mathbb{N}^{+}$ and $n_1 < n_2$. 
Let $\mathcal{H}_{n_1}$ be the Hilbert curve of order $n_1$ and $\mathcal{H}_{n_2}$ be the Hilbert curve of order $n_2$. 
For example, in Figure~\ref{fig:hilbert_encoding}, we show the coarse grid of order 3 in Figure~\ref{fig:coarse} and the fine grid of order 4 in Figure~\ref{fig:fine}. 
Next, we use the Hilbert curve on the coarse grid of order $n_1$ to encode the low-frequency component, while the Hilbert curve on the fine grid of order $n_2$ to encode the high-frequency component, respectively. 
\begin{eqnarray}
E_{\text{low}} = \mathcal{H}_{n_1} (X_\text{low}), \quad 
E_{\text{high}} = \mathcal{H}_{n_2} (X_\text{high}).
\end{eqnarray}

After obtaining the partial embedding, we input the low- and high-frequency data into the foundation model $\mathcal{M}$. 



\subsection{Frequency Decomposed Twogrid Approach}

In this section, we outline the algorithm for training twogrid MAEs in Algorithm~\ref{alg:multigrid_mae}.
Specifically, with the frequency decomposition process described in Sec.~\ref{sec:meth_freq}, one can decompose the input seismogram into two components, high frequency and low frequency, respectively. 
Next, we used a single neural network to reconstruct the high- and low-frequency components, respectively. 
The model structure is the small size Transformer with 256 heads, 12 Transformer blocks encoder and 4 Transformer blocks decoder. 
Let $X_{\text{high}}$ represent the high-frequency component of the input seismogram image, while $X_{\text{low}}$ denote the decomposed low-frequency component. 
Let $\mathcal{M}$ be the trained foundation models. 
Accordingly, the loss function for the high-frequency model is
\begin{eqnarray}
\mathcal{L}_{\text{high}} 
= \| X_{\text{high}} - \mathcal{H}_{n_2}^{-1} \mathcal{M}(\mathcal{H}_{n_2}(X_{\text{high}})) \|_2.
\end{eqnarray}

Similarly, the loss function for the low-frequency model is 
\begin{eqnarray}
\mathcal{L}_{\text{low}} 
= \| X_{\text{low}} - \mathcal{H}_{n_1}^{-1} \mathcal{M}(\mathcal{H}_{n_1}(X_{\text{low}})) \|_2.
\end{eqnarray}

The reconstructed loss is the weighted sum of high-frequency and low frequency $\ell_2$ loss, i.e., 
\begin{equation}
\mathcal{L}_{\rm{twogrid}}  = \alpha \mathcal{L}_{\rm{high}} + 
(1 - \alpha) \mathcal{L}_{\rm{low}},
\end{equation}
where $\alpha \in [0,1]$. 
If we use $\alpha = 0$, we learn only with the high-frequency components. For $\alpha =1$, the model is trained solely on low-frequency components. 

\subsection{Adaptive Frequency Decomposed Twogrid Approach}
\label{sec:ada_freq}
%

Motivated by the frequency principle~\cite{rahaman2019spectral,xu2019training,xu2019frequency}, a deep neural network (DNN) tends to learn a target function starting at low frequencies and progressing to high frequencies during training.
In this section, we propose an adaptive training strategy specifically for training two-grid masked auto-encoder (MAE) models. 
The loss function is defined as follows:
\begin{eqnarray}
\label{eq:adap_loss}
\mathcal{L}_{\text{adap}} = \alpha_t \mathcal{L}_{\text{high}} + (1 - \alpha_t) \mathcal{L}_{\text{low}},
\end{eqnarray}
where $\alpha_t = \alpha {t}/{T}$ is a linear decay function that depends on the ratio of the current iteration step to the total number of training steps, ${t}/{T}$. Consequently, $1 - \alpha_t$ represents a linear increase.
According to this formulation, the neural network initially focuses more on the low-frequency components. As training progresses, it gradually shifts its attention to the high-frequency components, aligning with the frequency principle discussed earlier.
In conclusion, we summarize \modelname~methods in Algorithm~\ref{alg:multigrid_mae}.




\begin{algorithm}[t!]
\caption{\modelname:{A}daptive {F}requency-{D}ecomposed {T}wo-{G}rid Masked Auto-Encoder Training}
\label{alg:multigrid_mae}
\begin{algorithmic}[1]
\REQUIRE Seismogram image set $\mathcal{I}$, learning rate $\eta$, epochs $E$, batch-size $B$ \\
\REQUIRE Coarse/fine Hilbert orders $n_1$, $n_2$, spectral threshold $k_0$
\ENSURE Pretrained foundation model $\mathcal{M}$

\STATE Initialize $\mathcal{H}_{n_1}$, $\mathcal{H}_{n_2}$ \COMMENT{Coarse/fine Hilbert curves}

\FOR{each epoch $e \in \{0,\dots,E-1\}$}
    \STATE $\mathcal{B} \gets \text{Sample}(\mathcal{I}, B)$ \COMMENT{Random minibatch}
    
    \STATE $X_{\text{high}}, X_{\text{low}} \gets \text{FreqDecomp}(\mathcal{B}, k_0)$ \COMMENT{Eq.~\eqref{eq:fourier_transform}-\eqref{eq:inverse_fourier_transform}}
    
    \STATE $E_{\text{high}} \gets \mathcal{H}_{n_2}(X_{\text{high}})$ \COMMENT{Fine-grid encoding}
    \STATE $E_{\text{low}} \gets \mathcal{H}_{n_1}(X_{\text{low}})$ \COMMENT{Coarse-grid encoding}
    
    \STATE $\mathcal{L} \gets \text{AdaptiveMAELoss}(E_{\text{high}}, E_{\text{low}})$ \COMMENT{Eq.~\eqref{eq:adap_loss}}
    \STATE $\theta \gets \theta - \eta \nabla_\theta \mathcal{L}$ \COMMENT{Parameter update}
    
    \STATE $\hat{I}_{\text{high}}, \hat{I}_{\text{low}} \gets \text{Reconstruct}(E_{\text{high}}, E_{\text{low}})$
\ENDFOR
\end{algorithmic}
\end{algorithm}

\begin{figure*}[htbp]
\begin{subfigure}[b]{0.32\textwidth}
 \centering
 \includegraphics[width=\textwidth]{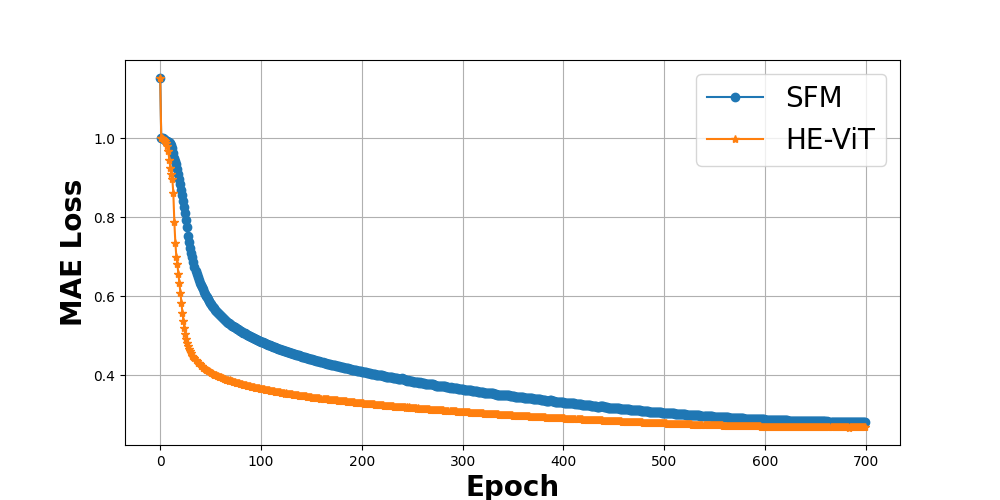}
\caption{Comparison of \base~and \hevit.}
\label{fig:hilbert_compare}
\end{subfigure}
 \begin{subfigure}[b]{0.32\textwidth}
     \centering
     \includegraphics[width=\textwidth]{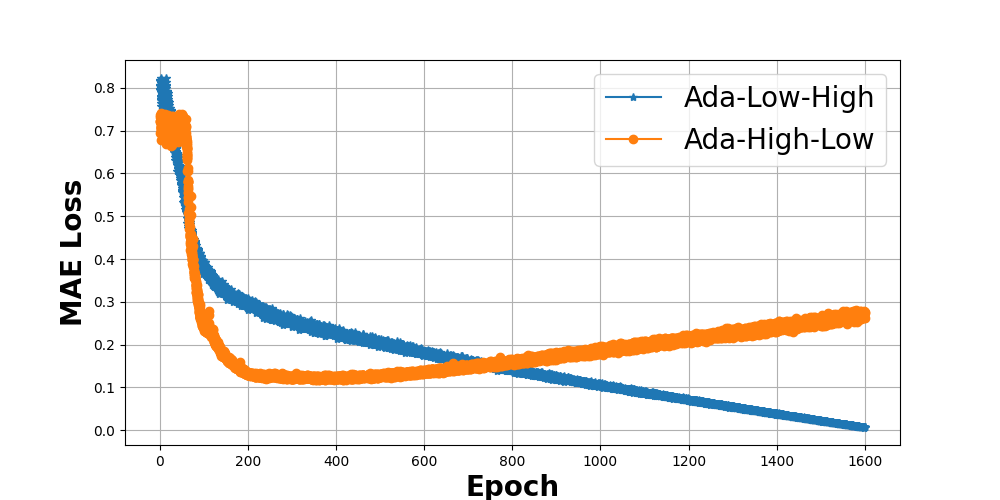}
    \caption{Frequency Principle for ViT.}
    \label{fig:freq_prin}
 \end{subfigure}
 \begin{subfigure}[b]{0.32\textwidth}
     \centering
     \includegraphics[width=\textwidth]{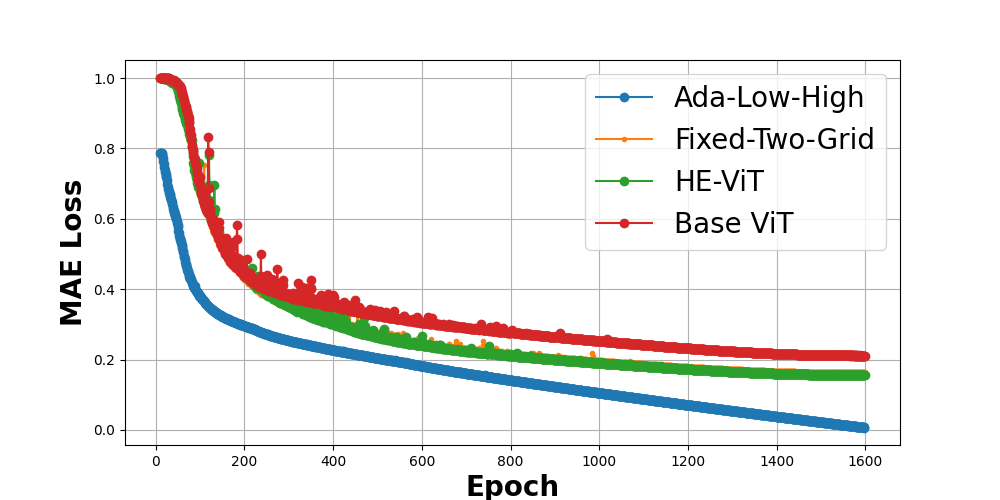}
    \caption{Improvement of Adaptive Strategy.}
    \label{fig:ada_improve}
 \end{subfigure}
 \caption{Training Dynamics.}
\end{figure*}

\section{Experimental Analysis}

In this section, we implement several experiments to answer the following research questions. 

\begin{itemize}
\item{\textbf{RQ.1}} How does Hilbert-ViT outperform the base version of ViT, and what specific improvements does it achieve?
\item{\textbf{RQ.2}} What is the choice for threshold of high- and low-frequency components? 
%
\item{\textbf{RQ.3}} What roles do high-frequency and low-frequency components play in the analysis and interpretation of the original input seismogram?
\item{\textbf{RQ.4}} How does the frequency principle inform and guide the development of an adaptive training strategy?
\item{\textbf{RQ.5}} What measurable improvements does the adaptive training strategy provide?
\end{itemize}

\subsection{Setup}

\subsubsection{Data Preparation}
This section outlines the data preparation process for training the foundation model, following the procedures described in SFM~\cite{sheng2025seismic}. The original three-dimensional seismic datasets are sourced from the United States Geological Survey (USGS)~\cite{usgs}, the South Australian Resources Information Gateway (SARIG)~\cite{sarig}, and the Society of Exploration Geophysics (SEG)~\cite{seg_open_data}. SFM~\cite{sheng2025seismic} performs two-dimensional seismic slicing in both inline and cross-line directions, resulting in over 2,200,000 segments, each measuring \(244 \times 244\) pixels. To enhance data diversity, we select 111,110 samples for training and 7,814 samples for testing. Additionally, we improve the input data resolution from \(224 \times 224\) to \(256 \times 256\) using reflection padding for the two-grid Hilbert encoding.



\subsubsection{Training Details}
We illustrate the model structure in Figure~\ref{fig:overall} (C), featuring an encoder with 12 Transformer blocks and a decoder with 4 Transformer blocks. The total parameter size is 84.57 MiB, with 81.82 MiB of trainable parameters.
Key hyperparameters for \modelname~include a batch size ($B$) of 336, a mask ratio of 0.75, 1,600 training epochs ($E$), a learning rate ($\eta$) of $1.5 \times 10^{-5}$, and a weight decay of 0.05.
Experiments were conducted on a server with four NVIDIA A100 GPUs, each having 80 GB of memory.


\begin{table*}[t]
\centering
\begin{tabular}{@{}l*{4}{c}@{}} 
\toprule
Methods & \base~\cite{sheng2025seismic} & \hevit & \twogrid & \rantwogrid\\
\midrule
MAE Loss $\downarrow$     & 0.1886 & 0.1535 & 0.1634 & 0.1665 \\
MSE $\downarrow$     & 1.6383 $\pm$ 0.3932 & 2.0629 $\pm$ 0.0965 & 1.8542 $\pm$ 0.0966 & 1.9132 $\pm$ 0.0791 \\
PSNR $\uparrow$      & 18.4969 $\pm$ 2.1720 & 15.7355 $\pm$ 1.7805 & 16.2393 $\pm$ 1.8034 & 13.4243 $\pm$ 1.3693 \\
SSIM $\uparrow$      & 0.1308 $\pm$ 0.0745 & 0.0620 $\pm$ 0.0425 & 0.0826 $\pm$ 0.0413 & 0.0620 $\pm$ 0.0404 \\
MS-SSIM $\uparrow$   & 0.2943 $\pm$ 0.2054 & 0.0731 $\pm$ 0.0693 & 0.0885 $\pm$ 0.0797 & 0.0327 $\pm$ 0.0510 \\
\midrule 
Methods &  \textbf{High} & \textbf{Low} & \modelone & \modeltwo  \\ 
MAE Loss $\downarrow$   &  0.0667 & 0.2701 & \textbf{0.0068} & 0.1291 \\
MSE $\downarrow$    & 1.6409 $\pm$ 0.2595 & 0.4583 $\pm$ 0.4349 & \textbf{0.3988 $\pm$ 0.1544} & 1.4363 $\pm$ 0.3771 \\
PSNR $\uparrow$   & 6.2952 $\pm$ 2.6248 & 16.6024 $\pm$ 1.6867 & \textbf{27.3535 $\pm$ 2.890} & 19.9746 $\pm$ 2.1178 \\
SSIM $\uparrow$   & 0.0807 $\pm$ 0.0567 & 0.0286 $\pm$ 0.0149 & \textbf{0.7362 $\pm$ 0.0540} & 0.2640 $\pm$ 0.0782 \\
MS-SSIM $\uparrow$  & 0.0441 $\pm$ 0.0537 & 0.0471 $\pm$ 0.0645 & \textbf{0.3890 $\pm$ 0.3582} & 0.3131 $\pm$ 0.2605 \\
\bottomrule
\end{tabular}
\caption{Overall performance of different foundation models pre-training.}
\label{tab:overall}
\end{table*}

\subsection{Improvement by Hilbert Encoding (RQ.1)}

In this section, we show the improvement of the Hilbert encoding compared to the foundational ViT model (SFM) in the pre-training stage of the foundation model in Figure~\ref{fig:hilbert_compare}. 
Observed in Figure~\ref{fig:hilbert_compare}, we observe that the Hilbert encoding accelerates the training process with a faster convergence and arrives loss around 0.3104 within the first 200 epochs. 
As training progresses, the Base ViT arrives with a similar performance as the Hilbert Curve ViT and finally reaches the similar convergence point at epoch 700. 
The final loss of Base ViT at epoch 700 is 0.2817 , while the final loss of the Hilbert curve ViT at the same epoch is 0.2673. 
With Hilbert Encoding, one gains the loss descend of 0.0144. 
However, although based on Hilbert encoding ViT, we obtain faster convergence and smaller loss compared to the Base ViT, the Hilbert Curve ViT fails to improve the metrics such as MSE, PSNR, and SSIM for the test samples. Thus, we discuss the two-grid approach for better Hilbert Encoding.

\subsection{Spectral Decomposition Threshold (RQ.2)}
 By spectral Decomposition described in Sect.~\ref{sec:meth_freq}, one can obtain both informative high- and low-frequency components for the seismogram. Furthermore, we record the $\ell_2$ norm for high- and low-frequency compounds in Table~\ref{tab:freq_decomp}, which confirms the observation. 
When the threshold is small, for example, $k_0 = 4$, the high frequency occupies the main energy. However, for a large threshold $k_0 = 64$, most energy is concentrated in the low frequency component. 
For $k_0 = 16$, the energy for low frequency is $165.75$ while for high frequency is $139.03$. The high- and low-frequency components show similar energy in this situation. Then we choose the frequency threshold as $k_0= 16$ through the experimental parts.
In the supplementary materials, we present comprehensive visualizations that demonstrate the distinct characteristics of high- and low-frequency components at multiple threshold values $k_0 \in \{4, 8, 16, 32, 64, 128\}$.

\begin{table}[t]
\setlength{\tabcolsep}{4pt} 
\centering
\begin{tabular}{lccccc}
\toprule
Threshold $k_0$ & 4 & 8 & 16 & 32 & 64 \\
\midrule 
Low Frequency  & 25.8 & 112.2 & 165.8 & 205.5 & 221.7 \\
High Frequency & 222.3 & 187.6 & 139.0 & 80.6 & 29.3 \\
Original Image & 224.0 & 224.0 & 224.0 & 224.0 & 224.0 \\
\bottomrule
\end{tabular}
\caption{$\ell_2$-norm energy distribution of frequency components across splitting thresholds.}
\label{tab:freq_decomp}
\end{table}


\subsection{High and Low Frequency Components (RQ.3)}

\begin{figure*}
     \begin{subfigure}[b]{0.16\textwidth}
         \centering
         \includegraphics[width=\textwidth]{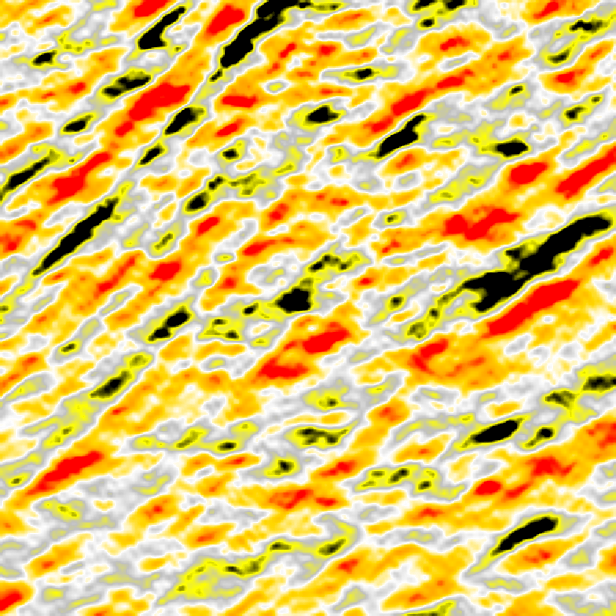}
         \caption{\textbf{Original}}
         \label{fig:original_decomp}
     \end{subfigure}
     \begin{subfigure}[b]{0.16\textwidth}
         \centering
         \includegraphics[width=\textwidth]{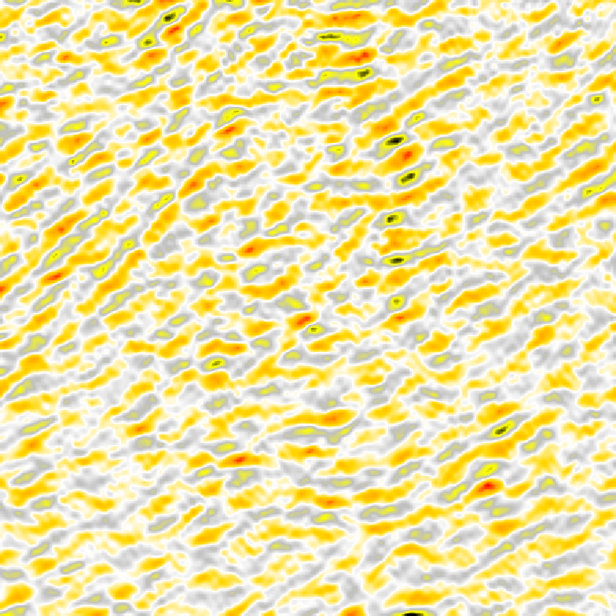}
         \caption{\textbf{High}}
         \label{fig:high_decomp}
     \end{subfigure}
     \begin{subfigure}[b]{0.16\textwidth}
         \centering
         \includegraphics[width=\textwidth]{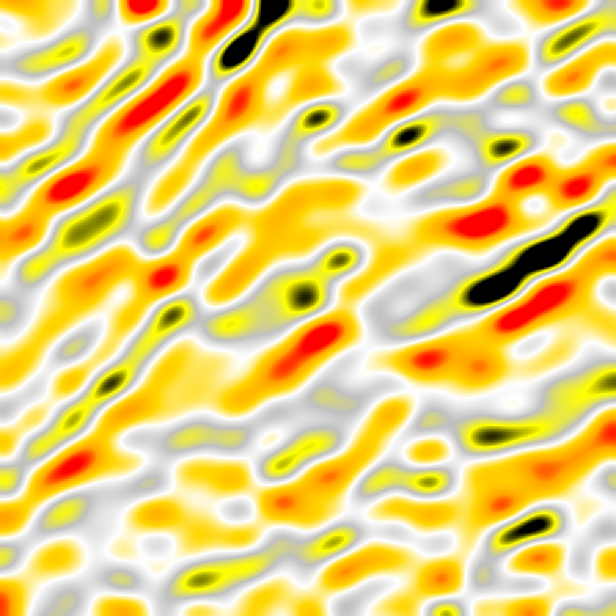}
         \caption{\textbf{Low}}
         \label{fig:low_decomp}
    \end{subfigure}
    \begin{subfigure}[b]{0.16\textwidth}
         \centering
         \includegraphics[width=\textwidth]{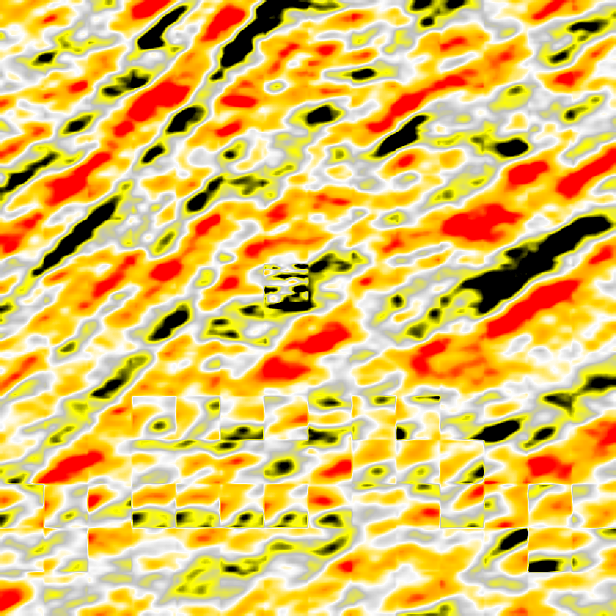}
         \caption{\textbf{Recons.}}
         \label{fig:original_recons}
     \end{subfigure}
     \begin{subfigure}[b]{0.16\textwidth}
         \centering
         \includegraphics[width=\textwidth]{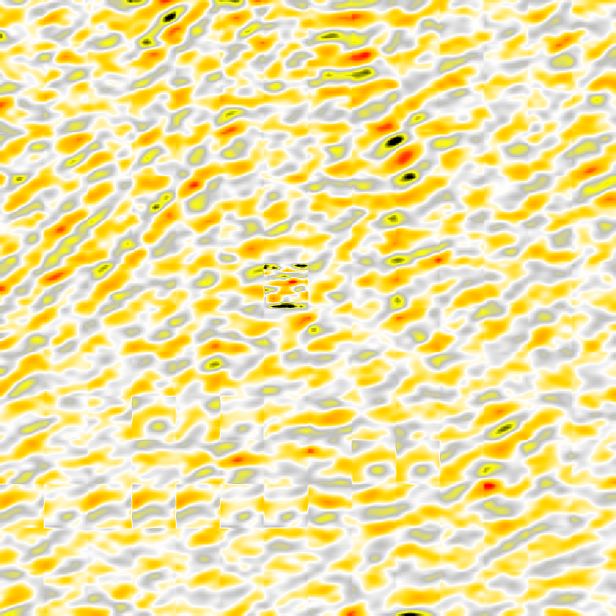}
         \caption{\textbf{\small{Recons.-High}}}
         \label{fig:high_recons}
     \end{subfigure}
     \begin{subfigure}[b]{0.16\textwidth}
         \centering
         \includegraphics[width=\textwidth]{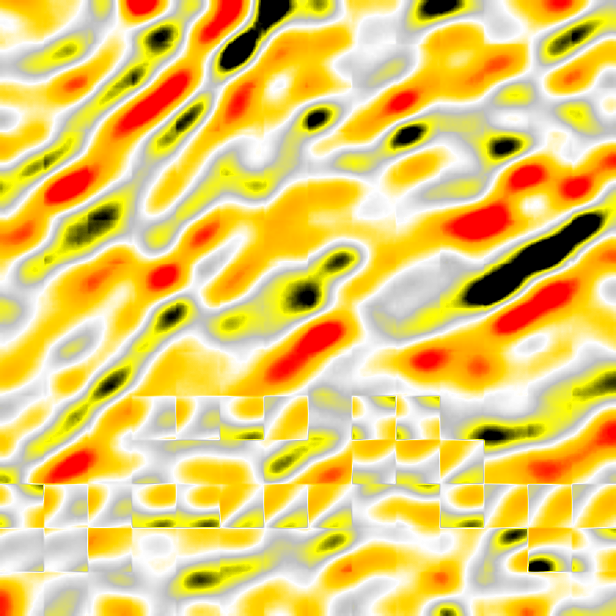}
         \caption{\textbf{\small{Recons.-Low}}}
         \label{fig:low_recons}
     \end{subfigure}
\caption{Decomposition and Reconstruction of High- and Low-Frequency Components.
The left part shows the original image (a), its high-frequency component (b), and its low-frequency component (c).
The right part displays the reconstructed image (d), with the reconstructed high-frequency (e) and low-frequency (f) components.
}
\label{fig:original_reconstruction}
\end{figure*}

In this section, we present the results for learning both high-frequency and low-frequency components. We utilize the fast Fourier transform to effectively separate these two components.
For our implementation, we choose a spectral decomposition threshold of 16. Independent reconstructions of the high-frequency and low-frequency models are illustrated in Figures \ref{fig:high_recons} and \ref{fig:low_recons}. 

The high-frequency components are associated with rapid changes in ground motion and correspond to shorter wavelengths. In contrast, low-frequency components relate to slower variations in ground motion and are linked to longer wavelengths.~\cite{yilmaz2001,sheriff1995}
High-frequency components are valuable for fine-structure identification, high-resolution imaging, and lithological analysis. Meanwhile, low-frequency components are useful for deep structure detection, velocity model construction, and full waveform inversion.~\cite{virieux2009,pratt1999}
By integrating high-frequency and low-frequency information, we can enhance multiscale imaging and improve the accuracy of our interpretations.~\cite{fichtner2010,tarantola2005}

\begin{table*}[t]
\setlength{\tabcolsep}{4pt} 
\centering
\begin{tabular}{lcccccc}
\toprule
Metrics & MAE Loss $\downarrow$ & MSE $\downarrow$ & PSNR $\uparrow$ & SSIM $\uparrow$ & MS-SSIM $\uparrow$ \\
\midrule
\modelone~(\textbf{Ada-Low-High}) & \textbf{0.0068} & \textbf{0.3988} & \textbf{27.3525} & \textbf{0.7362} & \textbf{0.3890} \\
\modelone~(\textbf{Ada-High-Low}) & 0.2769 & 4.1129 & 15.1371 & 0.0351 & 0.0454 \\
\modeltwo & 0.1291 & 1.4363 & 19.9747 & 0.2640 & 0.3131 \\
\bottomrule
\end{tabular}
\caption{Comparison of adaptive methods. Best results are indicated in bold.}
\label{tab:adap}
\end{table*}

\subsection{Frequency Principle (RQ.4)}
\label{sec:frequency_principle}
In this section, we experimentally verify the frequency principle for the Vision ViT architecture. We use the same structure as the ViT model for training \modelname, as shown in Fig.~\ref{fig:overall} part C. 
In contrast to \modelname, we include a comparative experiment that begins with training with the high-frequency components and then transitions to the low-frequency components at a linear rate. The loss function is defined as follows:
\begin{eqnarray}
\mathcal{L}_{\text{reverse}} = (1 - \alpha_t) \mathcal{L}_{\text{high}} + \alpha_t \mathcal{L}_{\text{low}},
\end{eqnarray}
where $\alpha_t = \alpha t/{T}$ serves as a linear decay coefficient for the high-frequency reconstructed loss, and $1 - \alpha_t$ indicates the linear increase coefficient for the low-frequency reconstructed loss. 

Figure~\ref{fig:freq_prin} illustrates the trend of loss using these two adaptive strategies. The strategy labeled as 'Ada-High-Low' refers to the approach where we initially focus entirely on high-frequency components and then gradually shift our attention to low-frequency components in a linear manner. Conversely, 'Ada-Low-High' describes the reverse strategy, where we start by concentrating on low-frequency components and then transition to high-frequency components gradually.
As shown in Figure~\ref{fig:freq_prin}, the 'Ada-Low-High' approach causes the loss to decrease significantly within a few epochs before increasing again after reaching its lowest point. In contrast, the 'Ada-High-Low' strategy decreases more slowly compared to 'Ada-Low-High' but continues to exhibit a decreasing trend even after reaching the end of training at 1,600 epochs.
The intersection between the two curves occurs around 800 epochs. 
In conclusion, the observations presented in Figure~\ref{fig:freq_prin} support the frequency principle for ViT blocks. Specifically, the model tends to learn low-frequency information at the beginning of training and gradually shifts to high-frequency information.

\subsection{Improvement of Adaptive Strategy (RQ.5)}

In this section, we discuss the improvements made to the adaptive strategy. Figure~\ref{fig:ada_improve} compares the performance of adaptive two-grid methods using the low to high strategy (Ada-Low-High) with base ViT, Hilbert curve ViT, and fixed two-grid methods, all evaluated over a long training period of 1,600 epochs.
In particular, the fixed two-grid ViT outperforms the Randomized Two-Grid methods. As shown in Figure~\ref{fig:ada_improve}, the base ViT achieves a loss of approximately 0.2114 after sufficient training epochs. The performance of the Hilbert Curve ViT and Fixed Two-Grid is quite similar, with final Mean Absolute Errors (MAE) of 0.1558 and 0.1634 respectively.
Using the adaptive training strategy, the MAE loss of Ada-Low-High is significantly reduced to around 0.0067, marking a decrease of up to 96.8\% compared to the base version. Furthermore, we observe that this method converges more quickly and has a lower initial MAE loss. The blue curves representing Ada-Low-High are consistently positioned below all other colored curves.
Moreover, we show the detailed comparison for \modeltwo, denoting the adaptive strategy with loss function(Eq.~\eqref{eq:adap_loss}), Hilbert encoding $\mathcal{H}_{n_2}$ for the high-frequency component, and plain encoding for the low-frequency component. Although \modeltwo~shows good visualization in Figure~\ref{fig:compare}, it performs worse than the evaluation metrics in Table~\ref{tab:adap}.
This may be induced by the visualization is mainly influenced via high-frequency information, while the low frequency plays a more important role in evaluation scores.


\section{Conclusion}
Driven by the intrinsic nature of seismogram data, which contains high- and low-frequency components, we developed a two-grid approach to enhance the training efficiency of our seismic foundation model. 
First, we decompose the input seismogram into high-frequency and low-frequency components. Next, we apply hierarchical Hilbert encoding to both components to preprocess the data for the ViT model. 
The training loss is calculated as a linear combination of the losses from the high-frequency and low-frequency components. We then enhance the two-grid training method by implementing an adaptive training loss construction. 
Finally, we conduct extensive experiments to demonstrate the effectiveness of the proposed methods. This approach can also be applied to various scenarios in computer vision where high- and low-frequency features differ significantly, yet both are essential.
In the future, we would expand \modelname~ to pre-train a seismic foundation model with the 3D seismogram as input and explore the improvements in the downstream task including full waveform inversion.

\section*{Acknowledgments}
This work is supported by the \textbf{Zhejiang Province Key Research and Development Plan} (Grant No.~2025SSYS0004), the \textbf{Major Research plan of the National Natural Science Foundation of China} (Grant No.~92370205), and 
the \textbf{National Natural Science Foundation of China} (Grant No.~62571529, and Grant No.~12271512).

\bibliographystyle{plain}
\bibliography{arxiv}

\appendix

In the technical appendix, we provide detailed descriptions of the model architecture in Sect.\ref{sec:detailed}, the two-grid Hilbert encoding methods in Sect.\ref{sec:twogrid}, the selection of spectrum splitting thresholds in Sect.\ref{sec:spec}, and the detailed construction of frequency principle in DNN in Sect.~\ref{sec:freq}. 
Lastly, we present the reconstruction performance across varying time steps in Sect.\ref{sec:recons_time}.

\begin{table*}[ht]
\centering
\caption{Architecture summary of the backbone model. The encoder consists of 12 identical transformer blocks, while the decoder includes 4 simplified blocks for masked token reconstruction.}
\begin{tabular}{p{4.5cm} p{10cm}}
\toprule
\textbf{Component} & \textbf{Configuration} \\
\midrule

\textbf{Patch Embedding} & 
Conv2d (in=1, out=768, kernel\_size=14, stride=14) \\

\textbf{Encoder (12$\times$ Blocks)} & 
Each block contains: \newline
\quad - LayerNorm (768) \newline
\quad - Multi-head Self-Attention: \newline
\quad\quad $\cdot$ Linear (768 $\rightarrow$ 2304) for QKV \newline
\quad\quad $\cdot$ Dropout + Linear (768 $\rightarrow$ 768) projection \newline
\quad - DropPath: Identity \newline
\quad - LayerNorm (768) \newline
\quad - MLP: \newline
\quad\quad $\cdot$ Linear (768 $\rightarrow$ 3072) $\rightarrow$ GELU $\rightarrow$ Linear (3072 $\rightarrow$ 768) \\

\textbf{Encoder Output Norm} & 
LayerNorm (768) \\

\textbf{Decoder Embedding} & 
Linear (768 $\rightarrow$ 256) \\

\textbf{Decoder (4$\times$ Blocks)} & 
Each block contains: \newline
\quad - LayerNorm (256) \newline
\quad - Multi-head Self-Attention: \newline
\quad\quad $\cdot$ Linear (256 $\rightarrow$ 768) for QKV \newline
\quad\quad $\cdot$ Dropout + Linear (256 $\rightarrow$ 256) projection \newline
\quad - DropPath: Identity \newline
\quad - LayerNorm (256) \newline
\quad - MLP: \newline
\quad\quad $\cdot$ Linear (256 $\rightarrow$ 1024) $\rightarrow$ GELU $\rightarrow$ Linear (1024 $\rightarrow$ 256) \\

\textbf{Decoder Output Norm} & 
LayerNorm (256) \\

\textbf{Prediction Head} & 
Linear (256 $\rightarrow$ 196) \\
\bottomrule
\end{tabular}
\label{tab:model_summary}
\end{table*}

\section{Detailed Model Architectures}
\label{sec:detailed}

In this section, we detail the model architecture corresponding to Figure 2(C) and summarized in Table~\ref{tab:model_summary}. The model adopts a Vision Transformer (ViT) backbone tailored for self-supervised representation learning via masked image reconstruction. It begins with a patch embedding layer that partitions the input image into non-overlapping patches using a convolutional projection with a kernel and stride size of 16. The encoder comprises 12 transformer blocks, each consisting of LayerNorm, multi-head self-attention with query-key-value (QKV) projections, dropout layers, and a two-layer MLP with GELU activation. The encoder output is then normalized and linearly projected into a lower-dimensional latent space, which serves as input to the decoder. The decoder consists of four lightweight transformer blocks, structurally similar to the encoder but operating with reduced dimensionality—256 hidden units and 1024 in the MLP. A final linear prediction head is used to reconstruct the masked image tokens. This design enables efficient contextual learning by encoding only the visible patches while predicting missing content, allowing the model to learn robust visual representations from sparse input. The complete model contains 88.68 million trainable parameters and occupies 84.57 MiB of storage.

\begin{figure*}[htbp]
    \centering
    \includegraphics[width=0.24\textwidth]{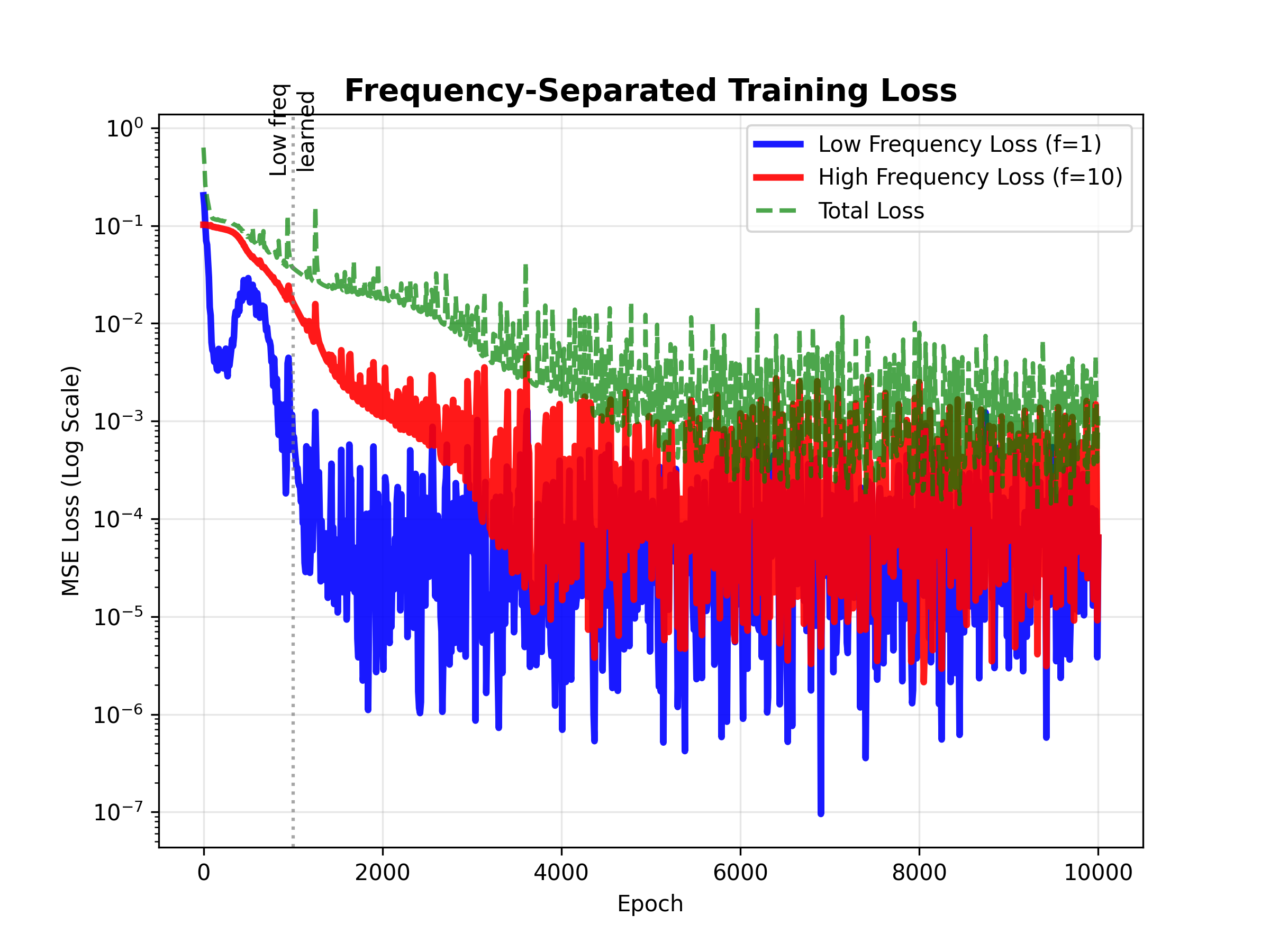}
    \includegraphics[width=0.24\textwidth]{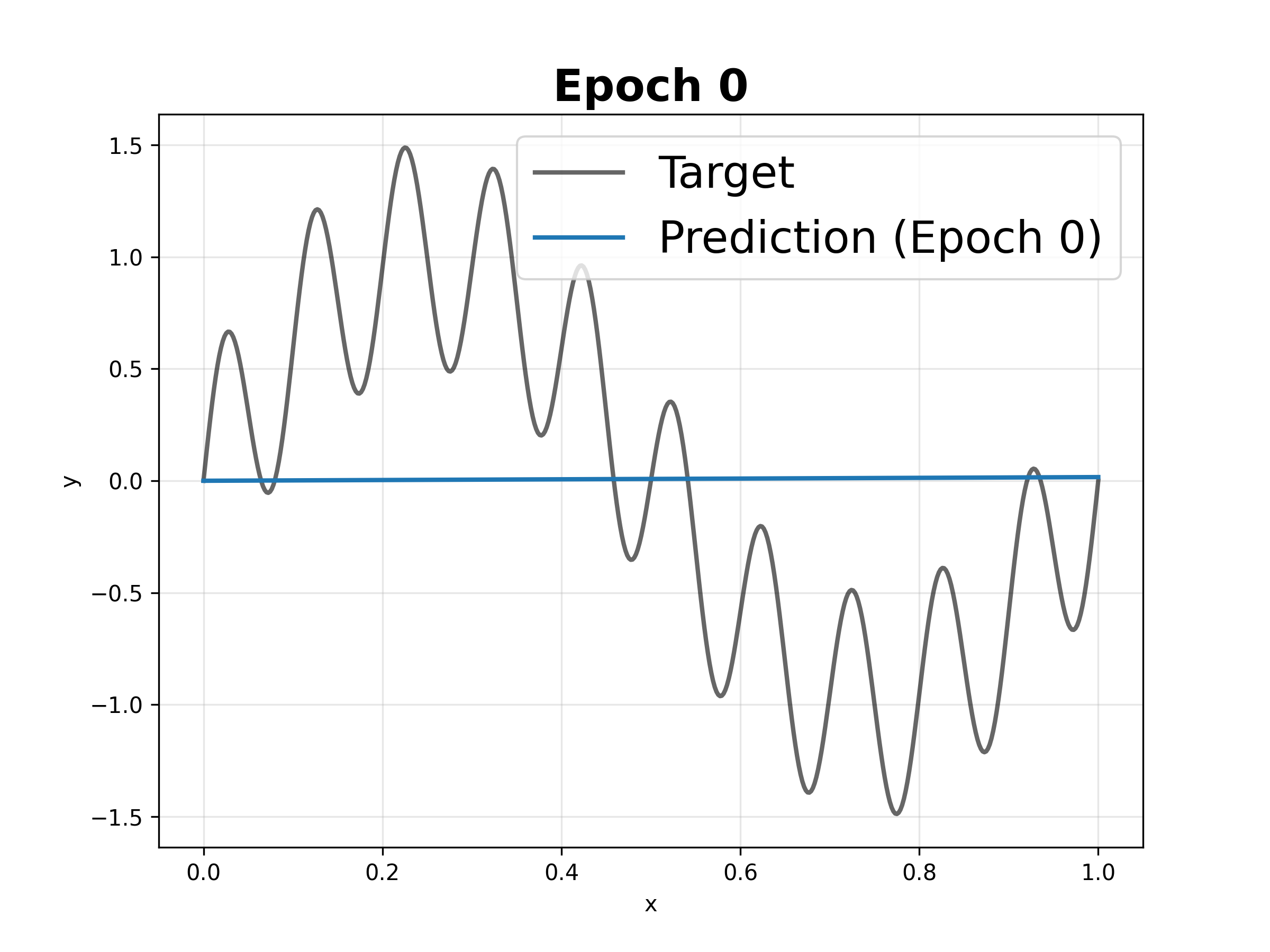}
     \includegraphics[width=0.24\textwidth]{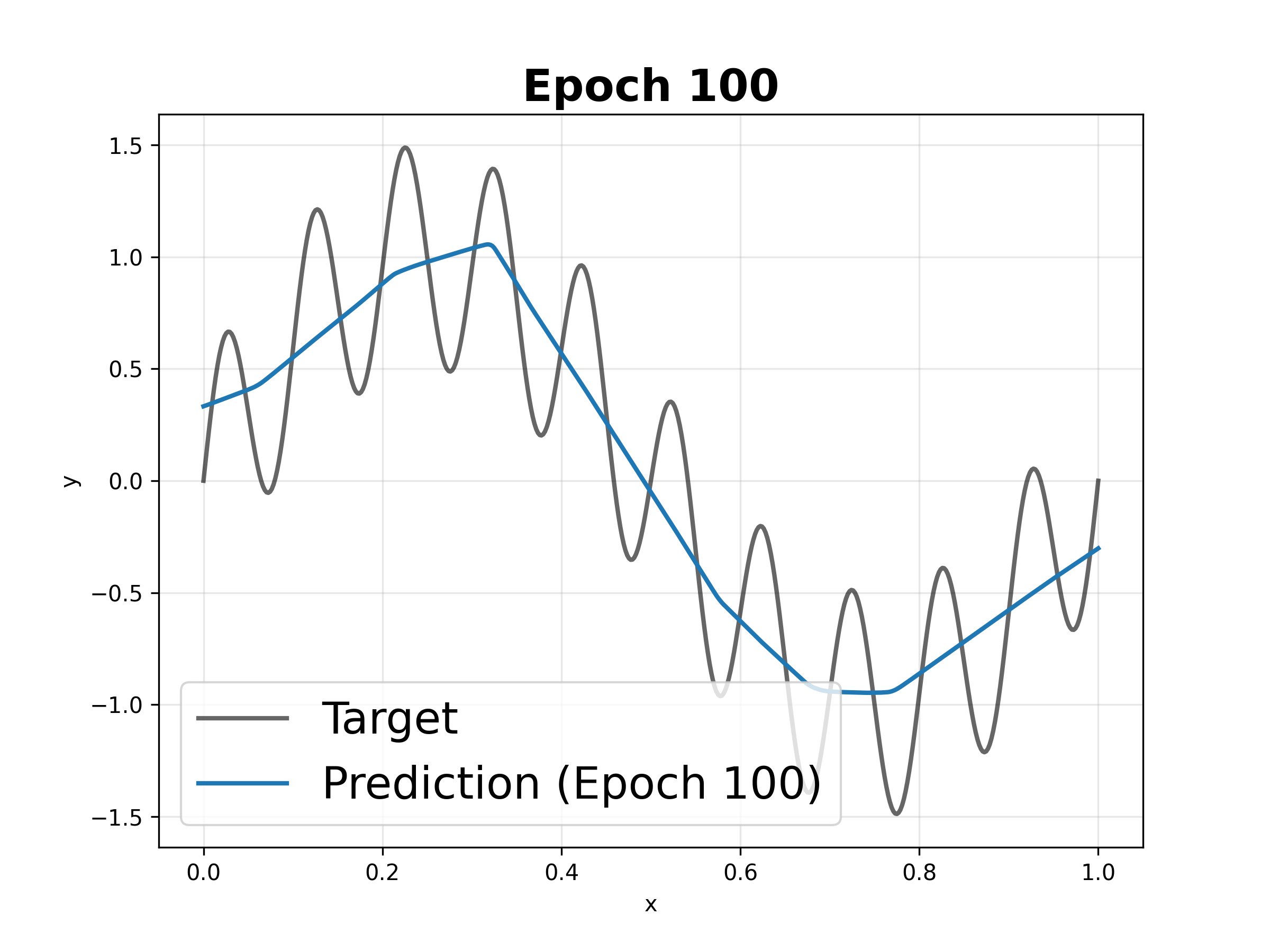}
    \includegraphics[width=0.24\textwidth]{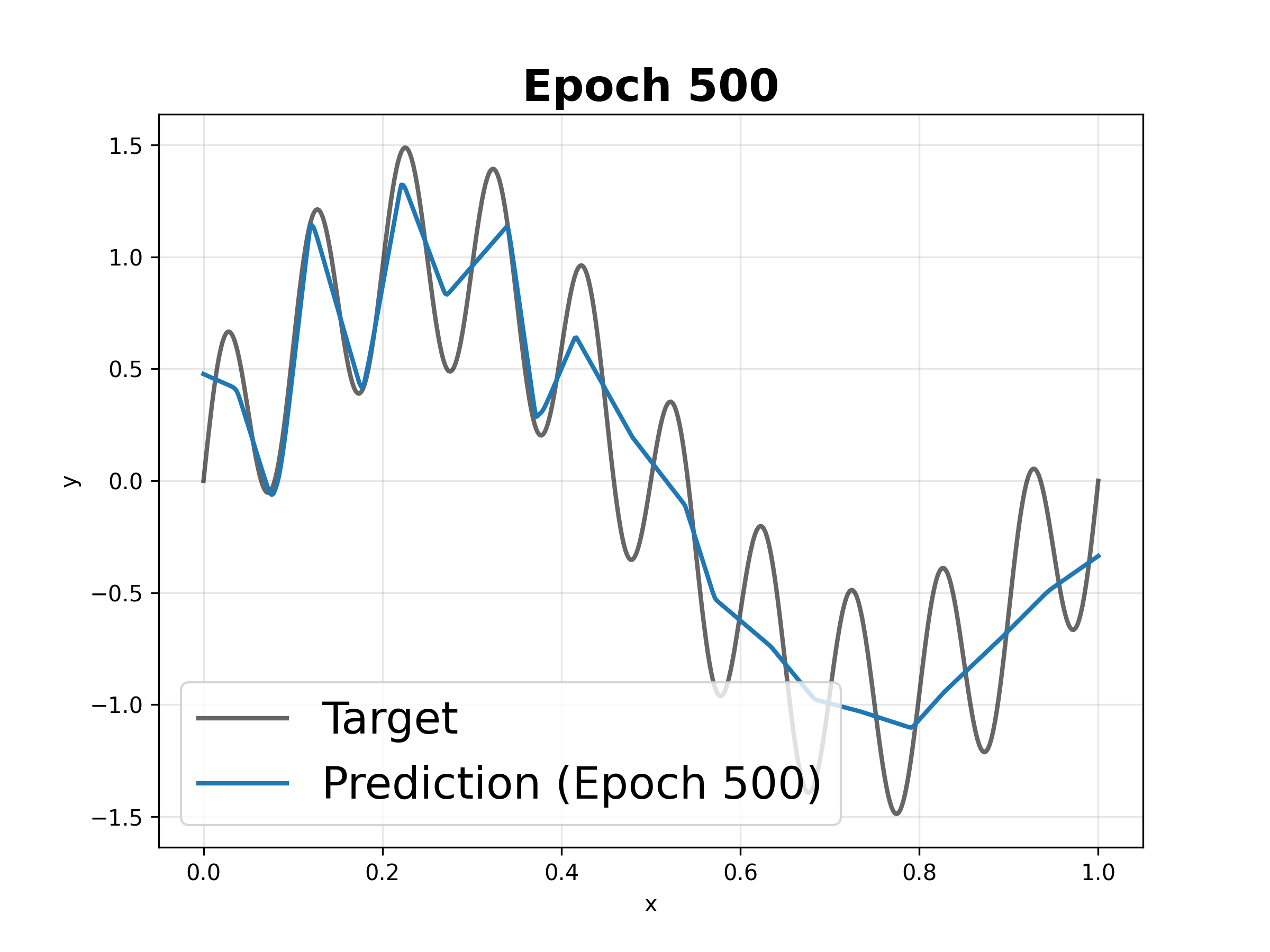}   
     \includegraphics[width=0.24\textwidth]{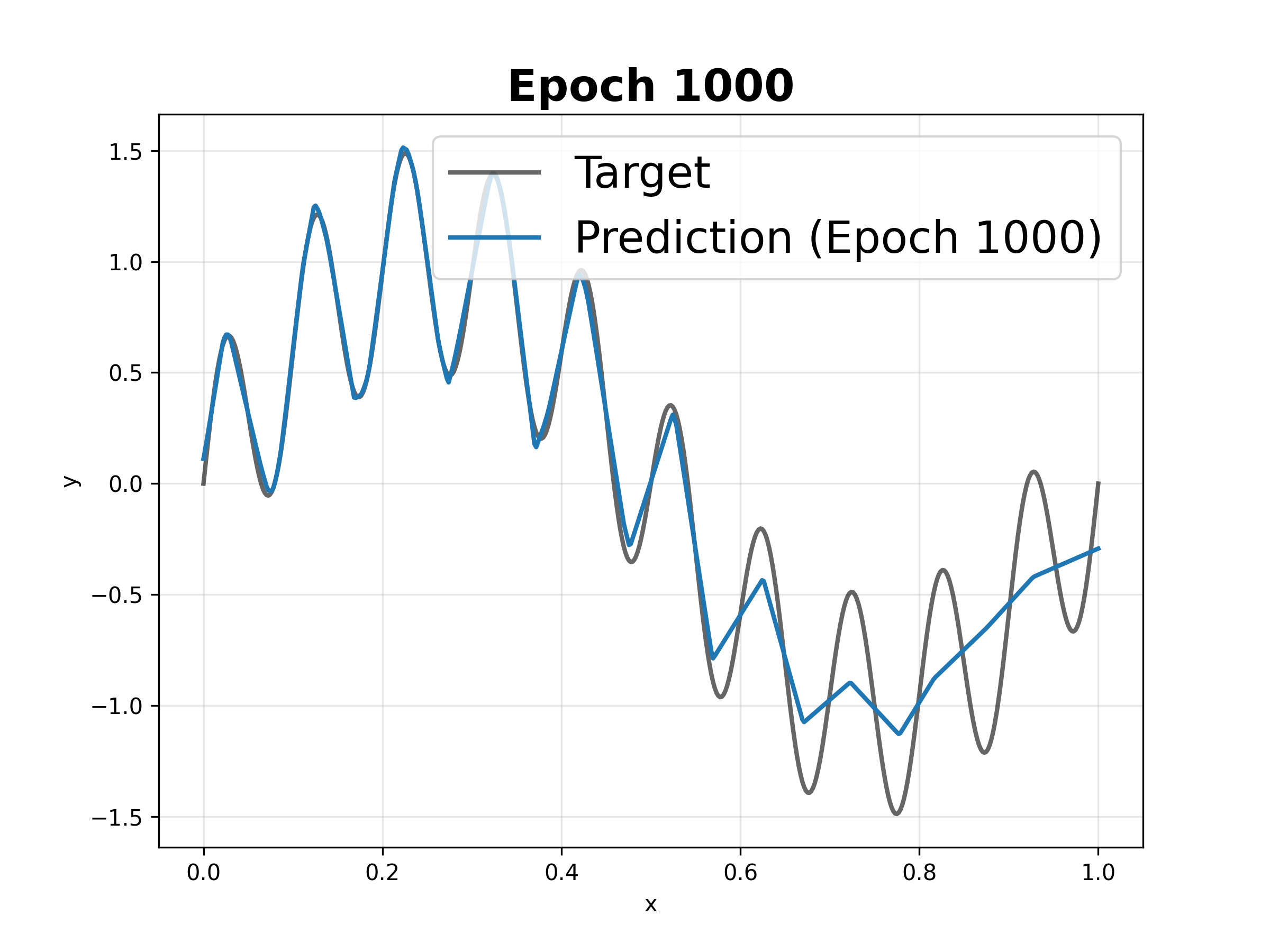}
    \includegraphics[width=0.24\textwidth]{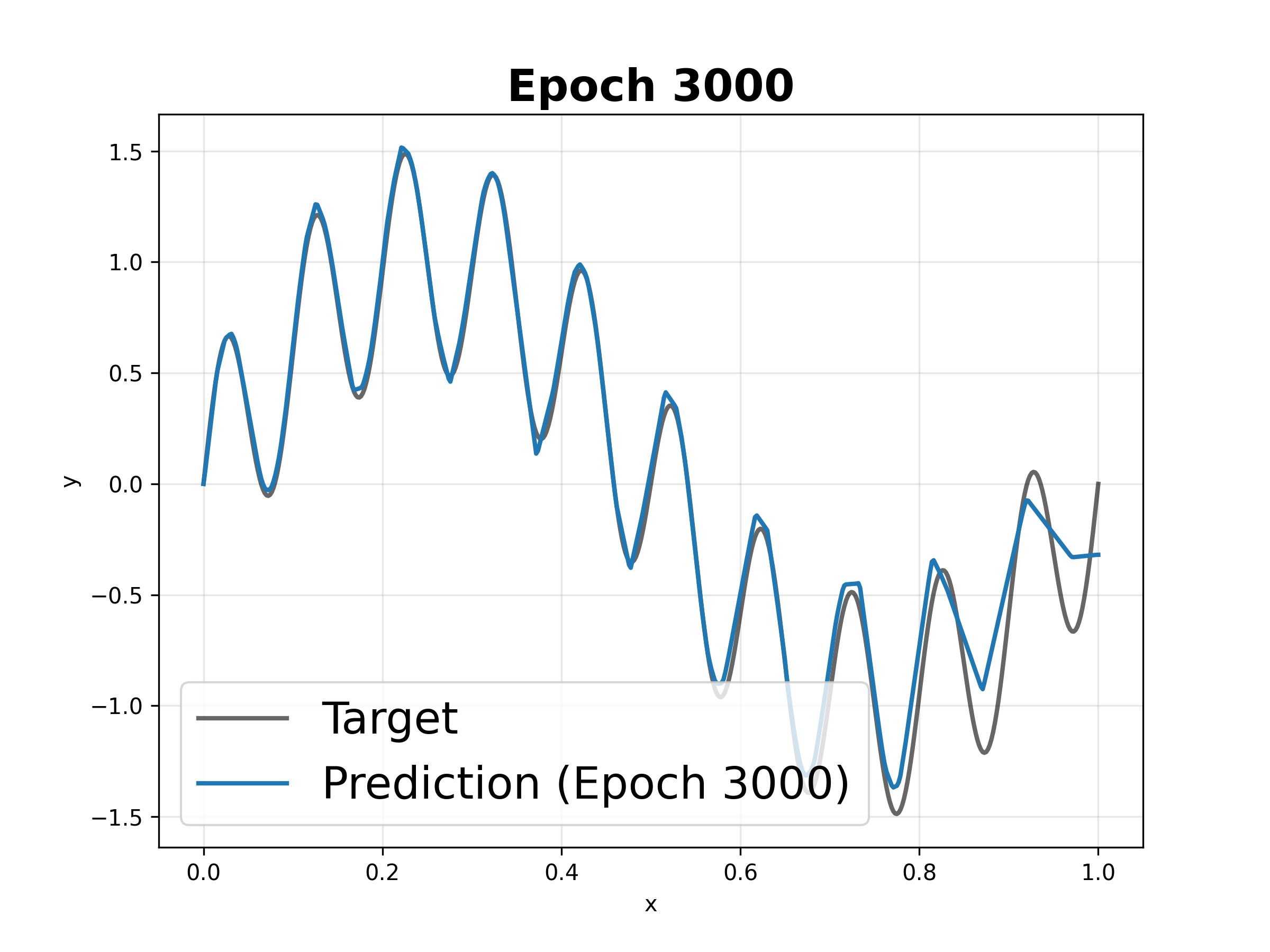}
    \includegraphics[width=0.24\textwidth]{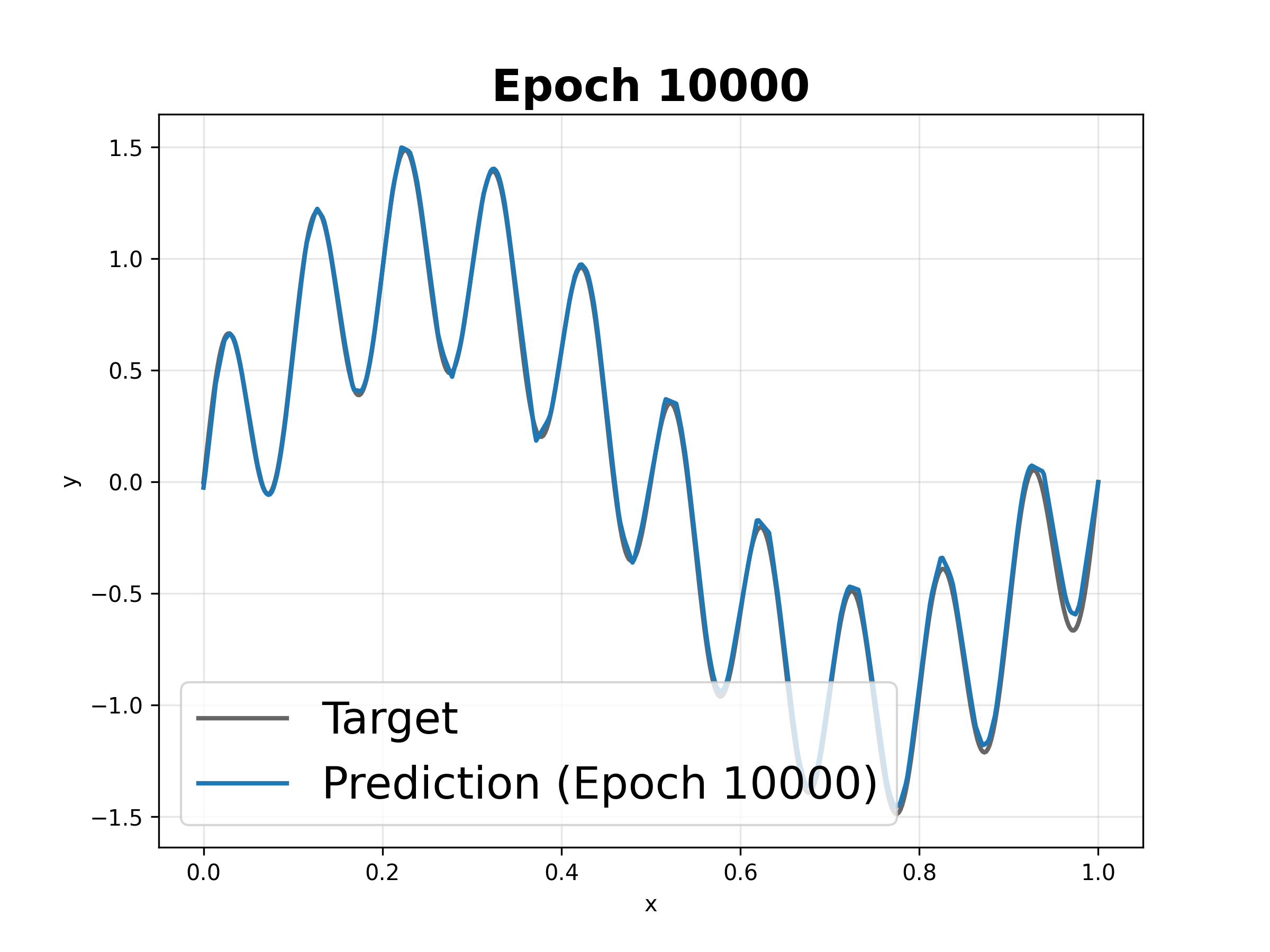}   
     \includegraphics[width=0.24\textwidth]{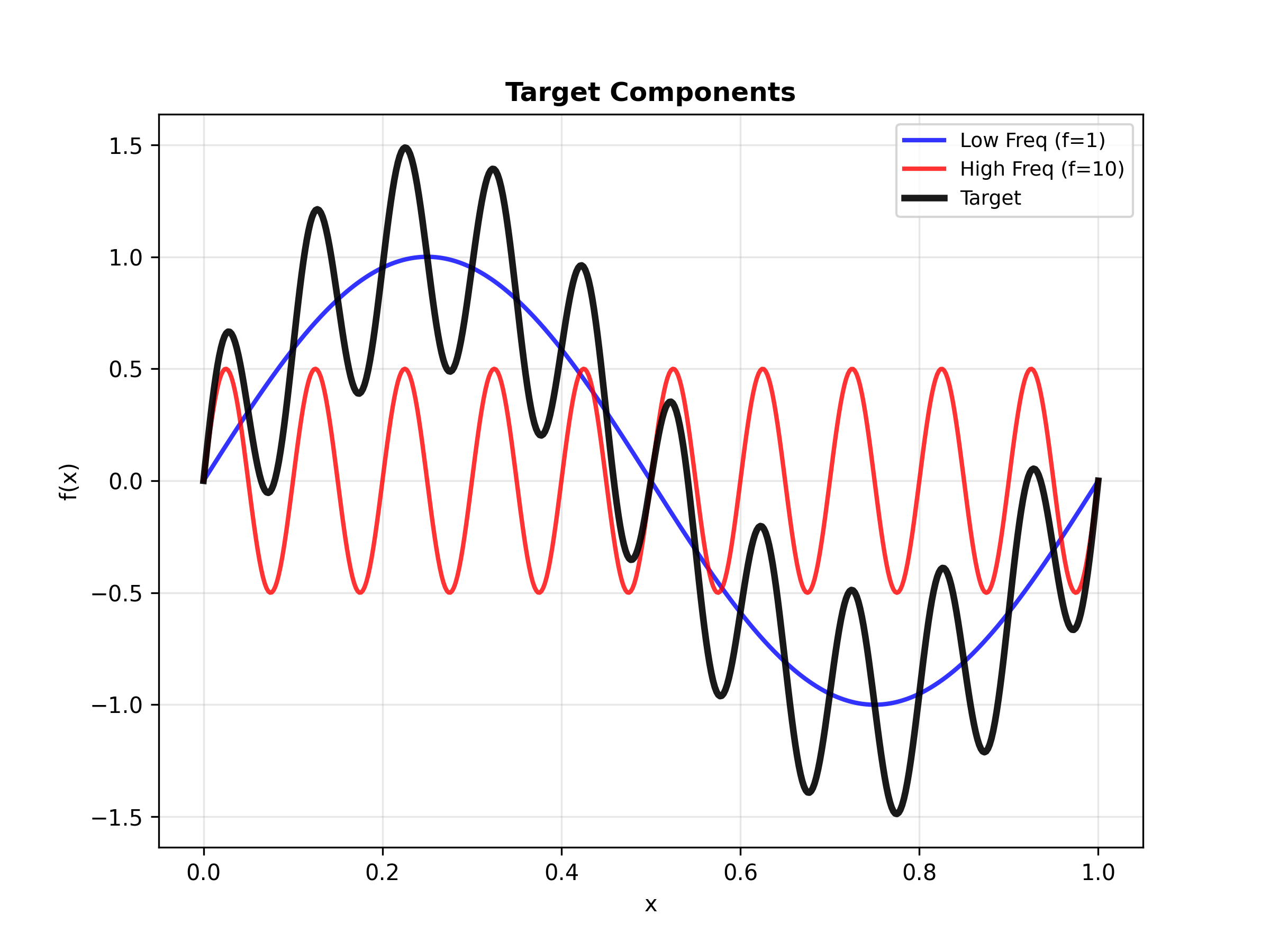}
     \includegraphics[width=0.24\textwidth]{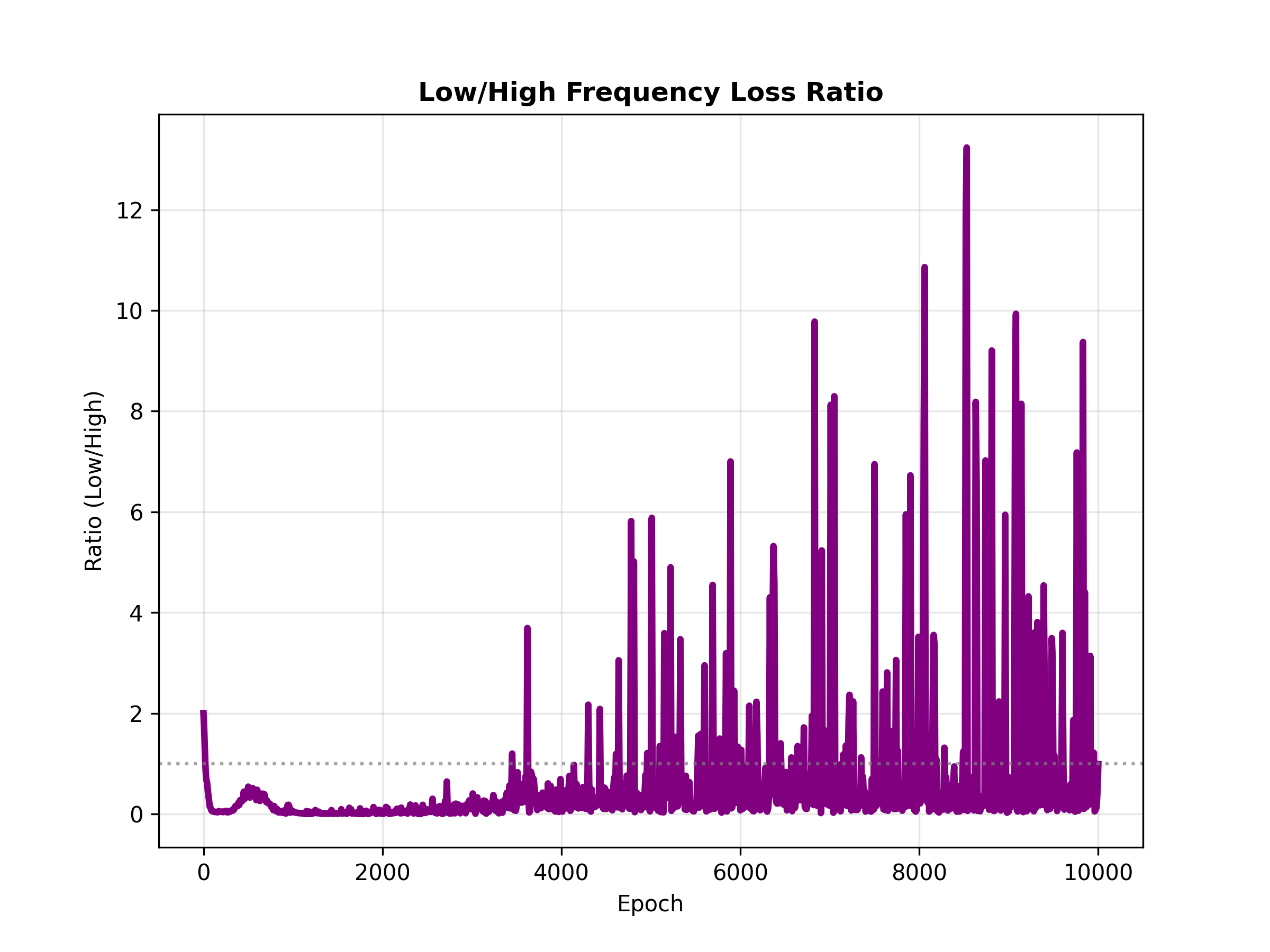}
    \includegraphics[width=0.24\textwidth]{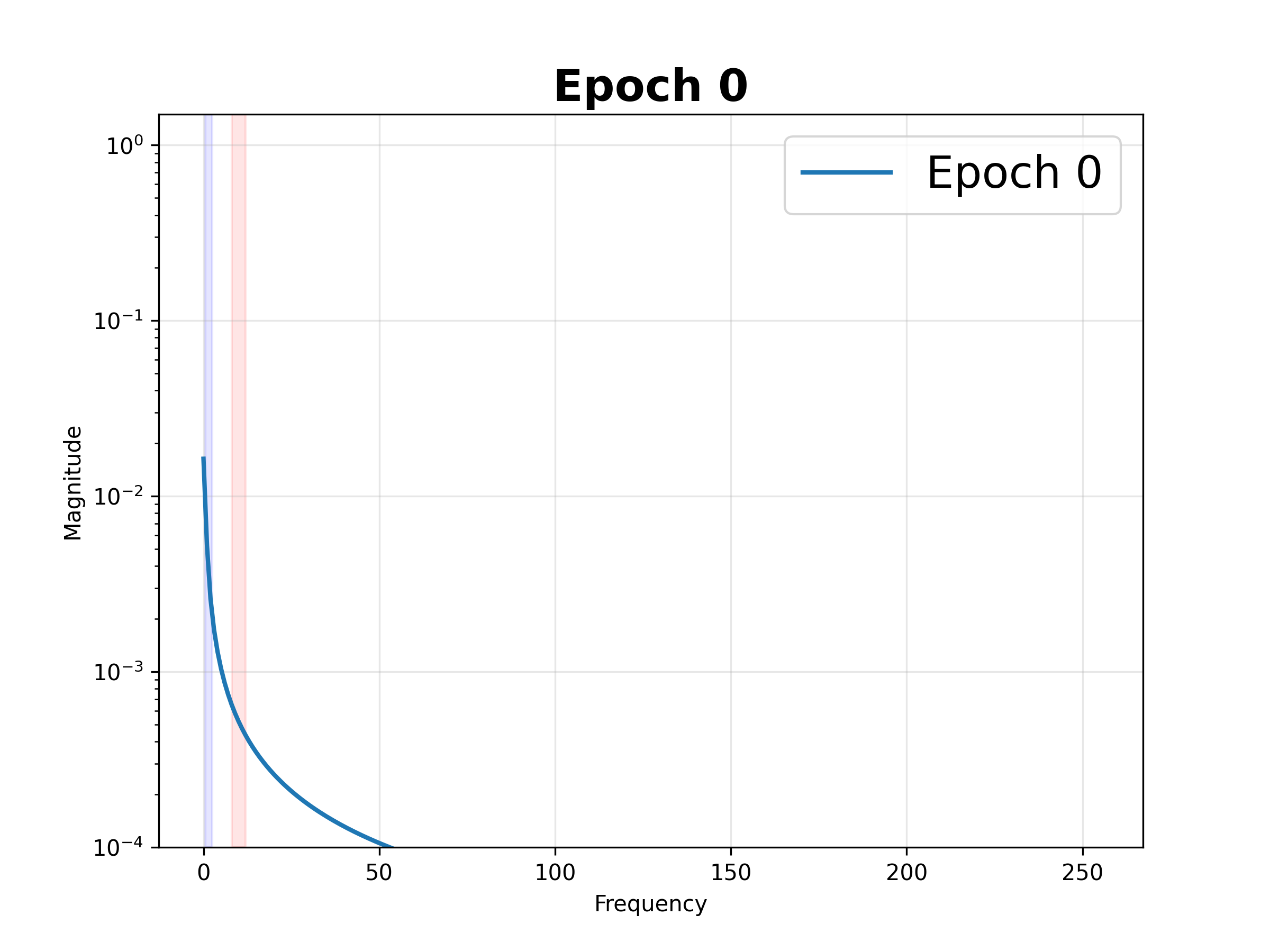}
    \includegraphics[width=0.24\textwidth]{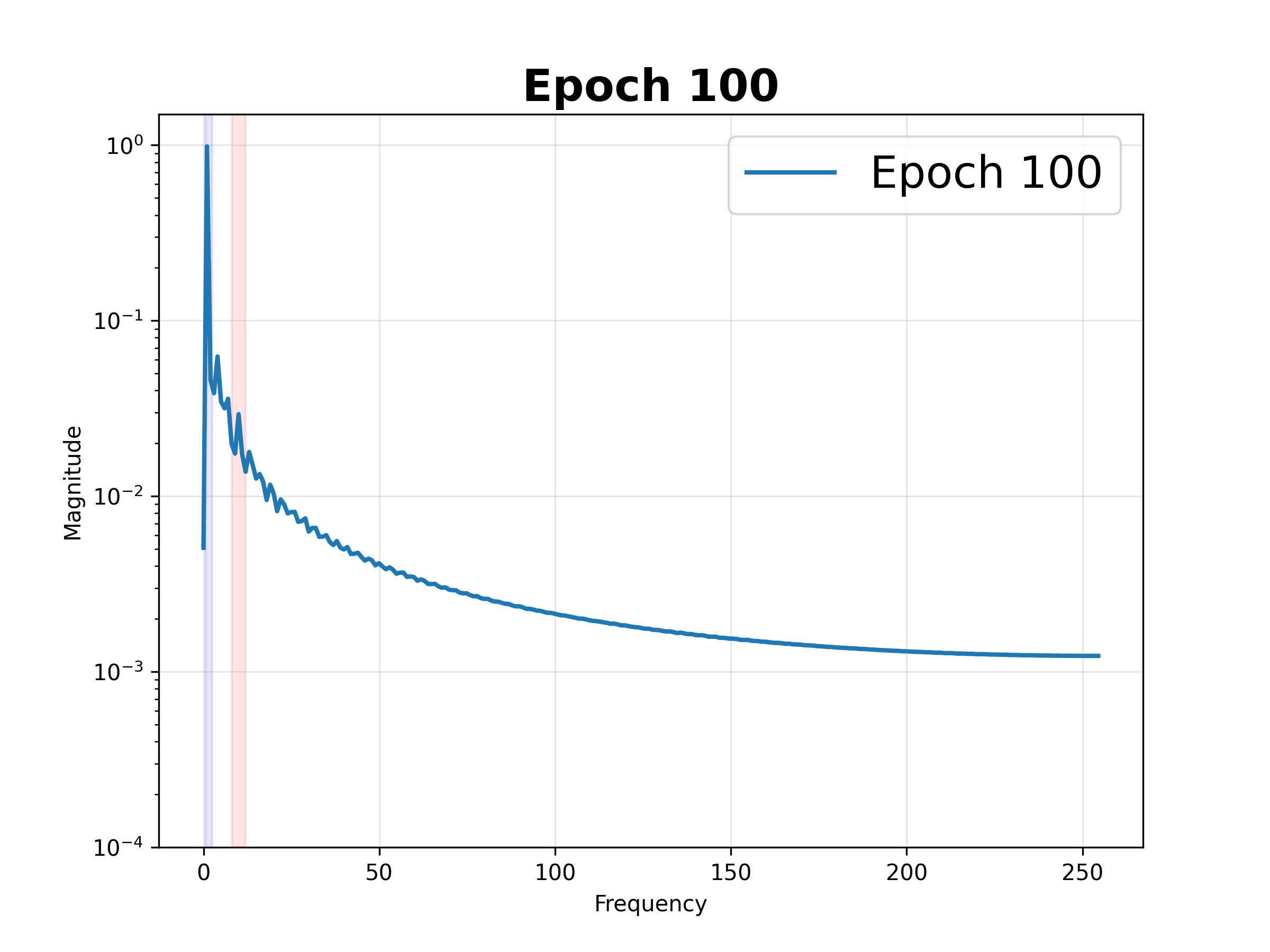}   
     \includegraphics[width=0.24\textwidth]{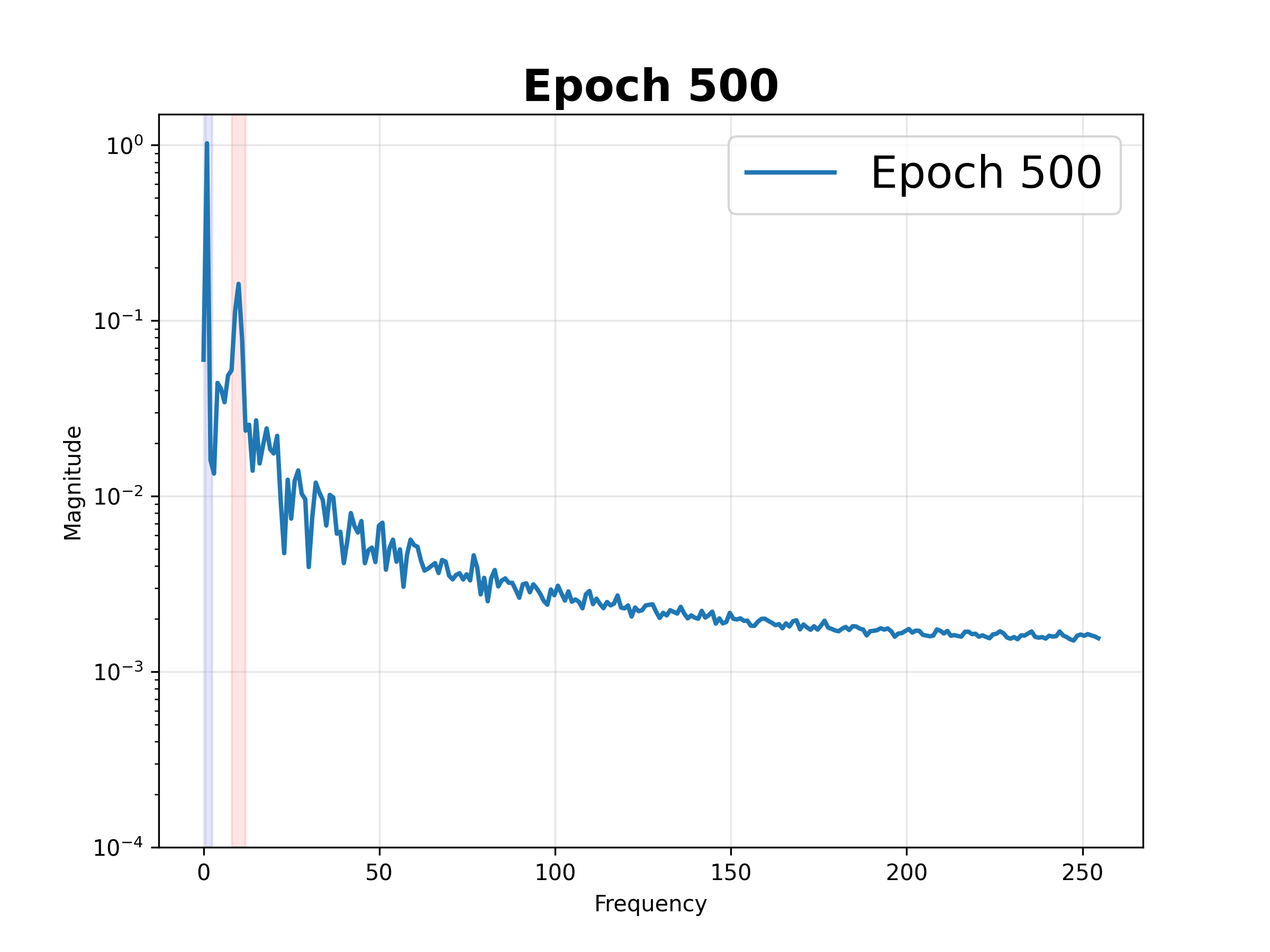}
     \includegraphics[width=0.24\textwidth]{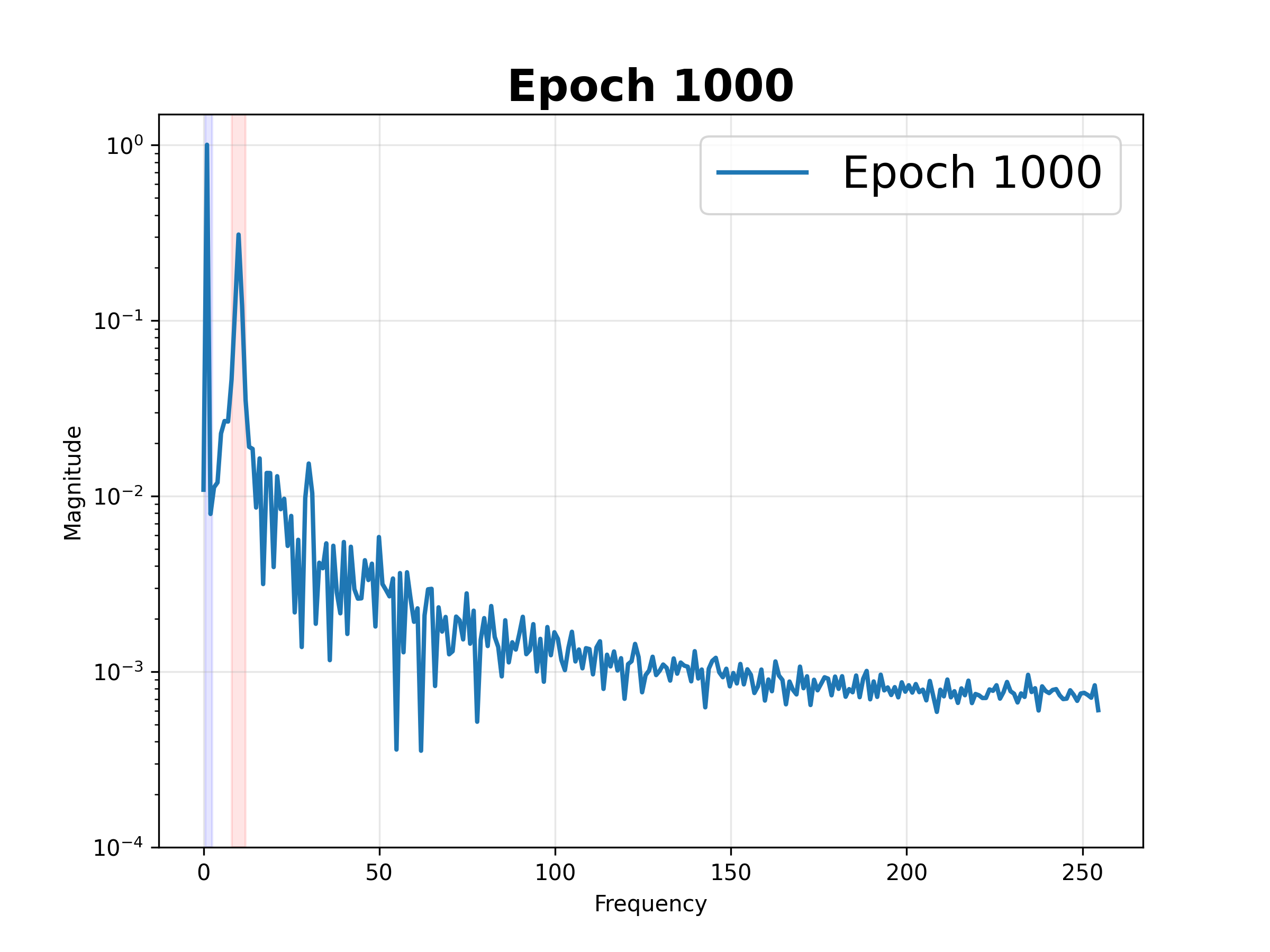}
    \includegraphics[width=0.24\textwidth]{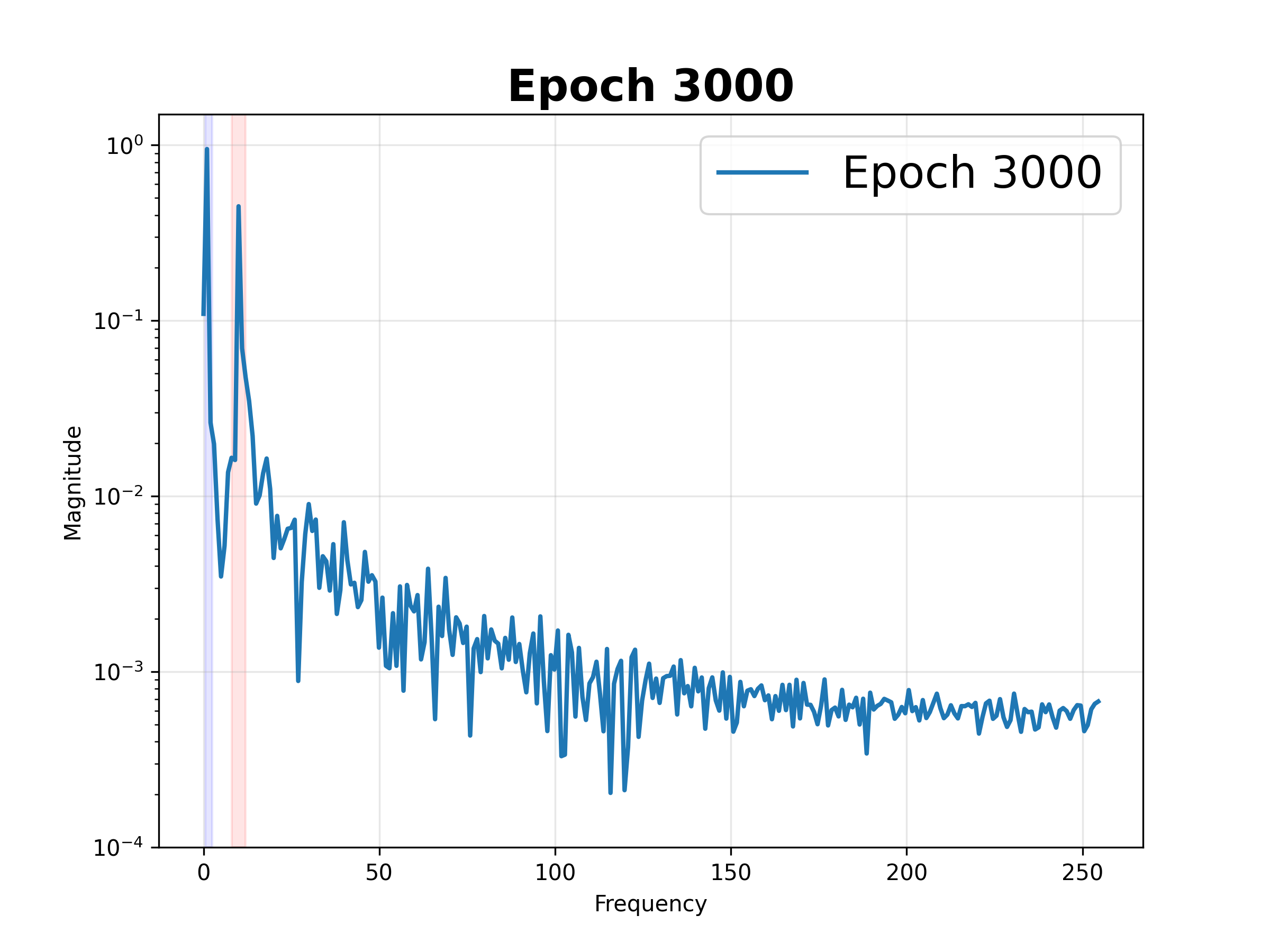}
    \includegraphics[width=0.24\textwidth]{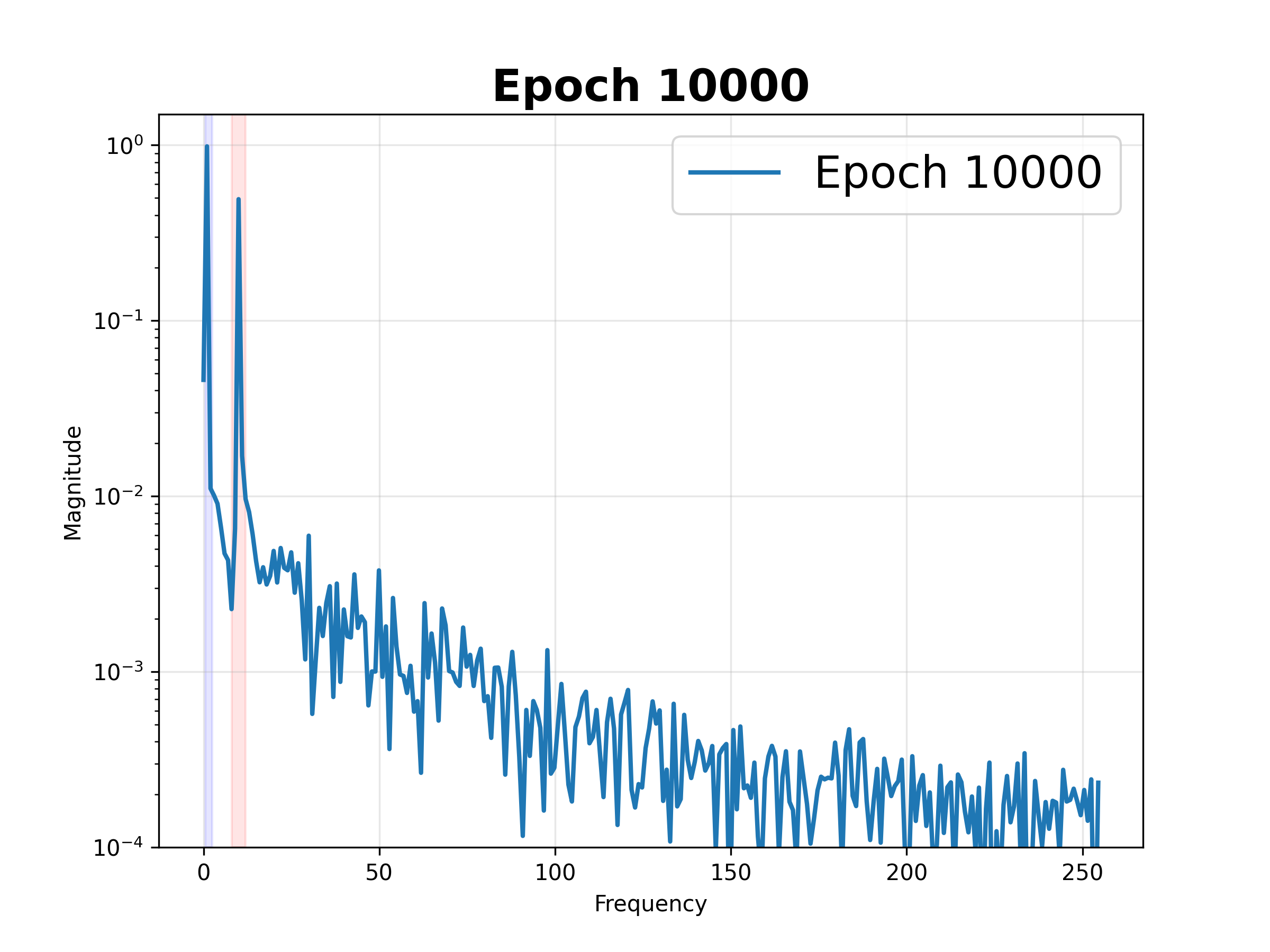}   
     \includegraphics[width=0.24\textwidth]{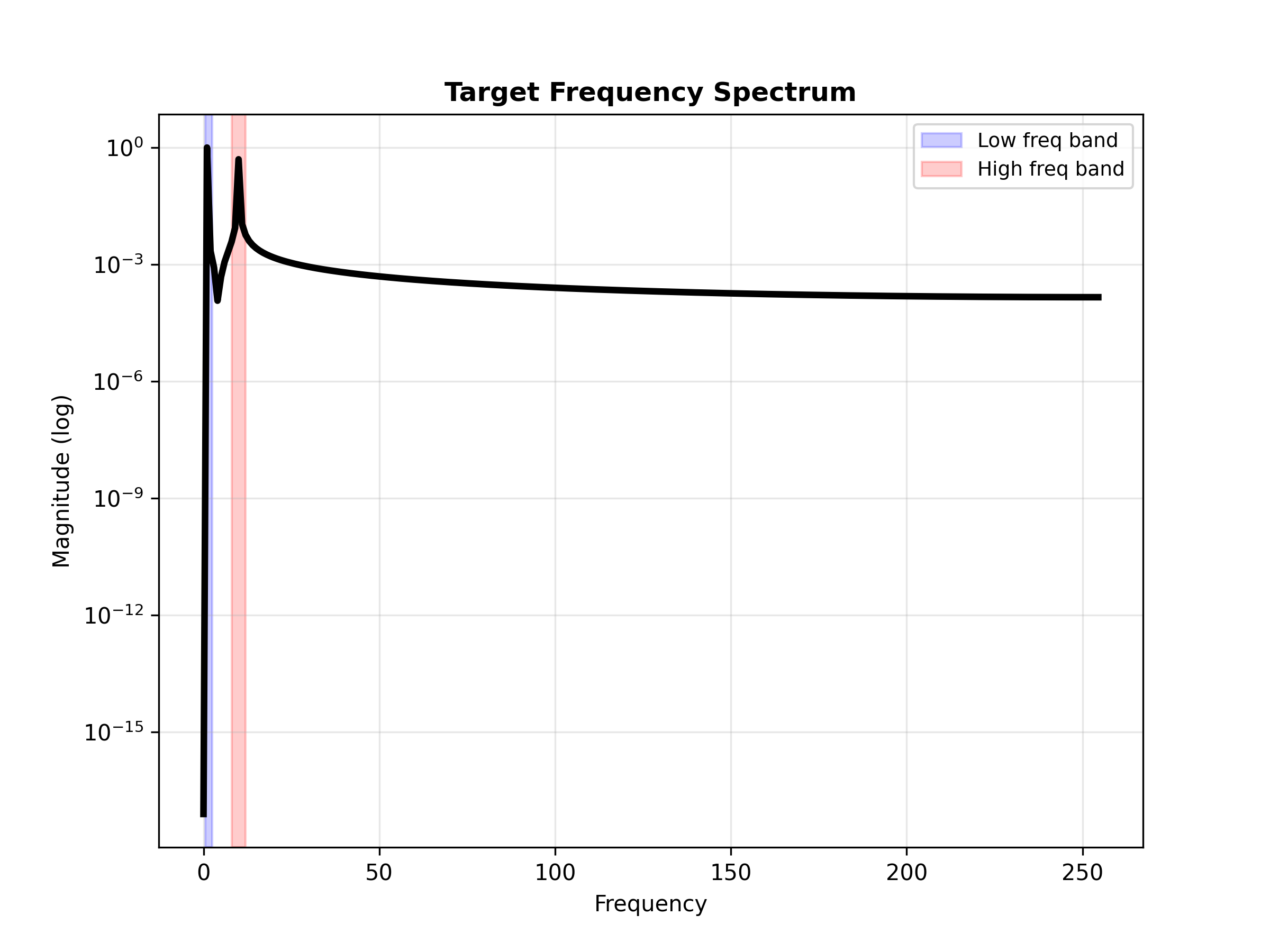}
    \caption{Frequency Principle Illustration Example.}
    \label{fig:freq_principle}
\end{figure*}

\section{Frequency Principle}
\label{sec:freq}

In this section, we present a detailed example of the frequency principle using a one-dimensional target function defined as:

$$
f(x) = f_1(x) + 0.5 f_2(x) = \sin(2 \pi x) + 0.5 \sin(20 \pi x),
$$

where \( f_1(x) = \sin(2 \pi x) \) represents the low-frequency components with a frequency of \(1\), while \( f_2(x) = \sin(20 \pi x) \) captures the high-frequency components with a frequency of \(10\).

We utilize a six-layer dense neural network (DNN) with a width of 200 and ReLU activation functions. The optimizer used for training is Adam~\cite{adam2014method} with a learning rate of \(10^{-3}\). The model is trained using mean squared error (MSE) as the loss function.

Training is conducted over a total of 10,001 epochs, with snapshots taken at the following intervals: \{0, 100, 500, 1000, 3000, 10000\}.

In Figure~\ref{fig:freq_principle} (Row 1, Col 1), we illustrate the separated training losses for high-frequency (red) and low-frequency (blue) components, respectively. We observe that the low-frequency components decay rapidly in the first 1,000 epochs and converge around 2,000 epochs. Conversely, the high-frequency components decay slowly during the initial 2,000 epochs and converge around 4,000 epochs. This behavior aligns with the claims of the frequency principle discussed in~\cite{xu2019frequency}.

The remaining figures in Rows 1 and 2 visualize the approximation effects of DNNs. Specifically, in the first 1,000 epochs, the DNN captures the low-frequency pattern information, which represents the coarse-level shape of the signal. In contrast, the high-frequency information is fitted around 3,000 epochs and becomes well-fitted by 10,000 epochs. The target signal, consisting of both high- and low-frequency components, is shown in Row 2, Column 4.

In Rows 3 and 4, we compute the spectrum distribution throughout the training process. In Row 3, Column 1, we present the high- and low-frequency loss ratio. In the remaining figures of Rows 3 and 4, we depict the spectrum distribution of the approximated functions. Notably, at epoch 0, the spectrum primarily concentrates on low frequencies. As training progresses, the spectrum expands to include high frequencies, ultimately converging on both the \(1\) and \(10\) frequency components. The final figure in Row 4, Column 4, illustrates the spectrum distribution of the exact function. This dynamic indicates that DNNs initially capture low frequencies before transitioning to higher frequencies. 

This observation serves as the design motivation for our adaptive training loss presented in Eq~(8), where we allow the model to first learn low-frequency information before gradually focusing on high-frequency components.

\begin{figure*}[htbp]
  \centering
   \includegraphics[width=1\linewidth]{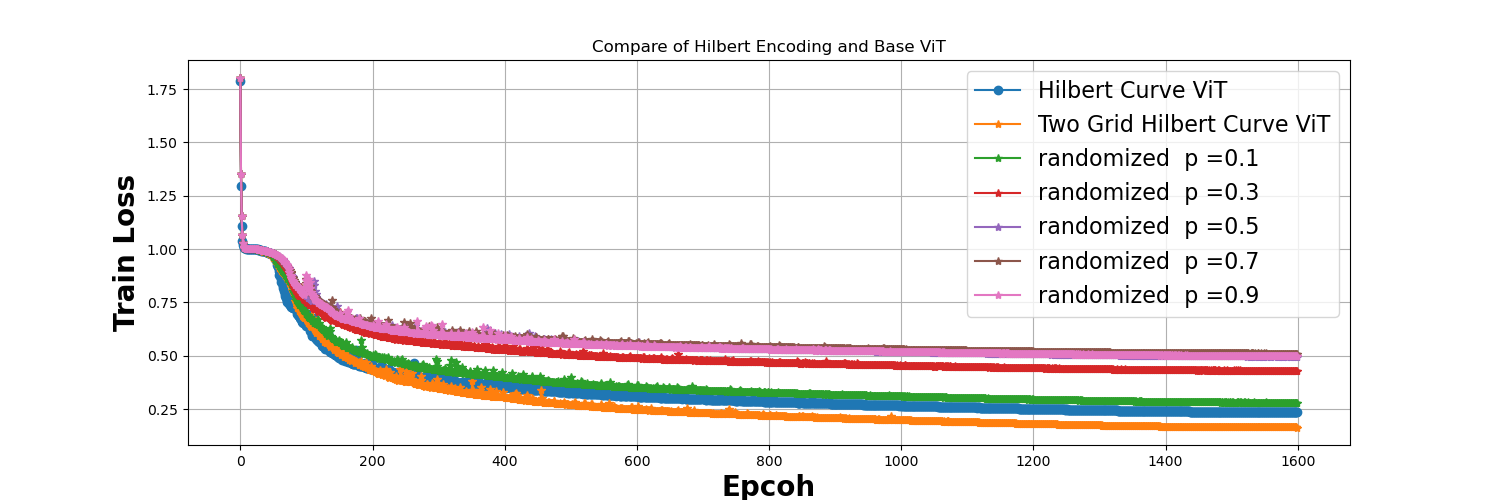}  
\caption{Comparison of \twogrid, \rantwogrid~with different sampling rates and \base. The high random probability the poor performance of \rantwogrid. The \twogrid~achieves best performance in this scenario.}
\label{fig:pure-mg}
\end{figure*}

\section{Two-Grid Hilbert Encoding (\twogrid~and~\rantwogrid)}
\label{sec:twogrid}

\begin{table}[t]
\centering
\begin{tabular}{cc}   
\toprule
Metrics & MAE Loss $\downarrow$  \\
\midrule 
\hevit & 0.2360 \\
\twogrid & \textbf{0.1634}  \\
\midrule 
\rantwogrid~ $p=0.01$ & \underline{0.1665}  \\
\rantwogrid~ $p=0.03$ & 0.1975  \\
\rantwogrid~ $p=0.05$ & 0.2189  \\
\rantwogrid~ $p=0.07$ & 0.2458  \\
\rantwogrid~ $p=0.09$ & 0.2680  \\ 
\bottomrule
\end{tabular}
\caption{Comparison of \base, \twogrid, and \rantwogrid~Methods.}
\label{tab:twogrid}
\end{table}

In this section, we present the experimental results of two-grid Hilbert encoding under both deterministic fixed-point and random settings. The first method refines the Hilbert encoding in a deterministic fixed-point manner, while the second method incorporates random refinement. During training, the points for the two-grid Hilbert encoding are selected randomly, with the probability of choosing the refined points set at $0.01, 0.03, 0.05, 0.07, 0.09$.
In Table~\ref{tab:twogrid}, we compare three methods: one-layer Hilbert encoding (referred to as \hevit), two-grid deterministic Hilbert encoding (denoted as \twogrid), and randomized two-grid methods with varying sampling probabilities (labeled as \rantwogrid).
From Table~\ref{tab:twogrid}, it can be seen that encoding with the TwoGrid Hilbert sequence significantly improves the mean absolute error (MAE) loss compared to the OneGrid Hilbert encoding, reducing the MAE loss from 0.2360 to 0.1634. 

The introduction of randomness leads to an increase in MAE loss. As the probability $p$ increases, the likelihood that any pixel will be selected for the second-order Hilbert encoding also increases, contributing to greater randomness in the training process. The second column in Table~\ref{tab:twogrid} shows that a higher probability results in an increased MAE loss, indicating a decline in the performance of the pre-trained foundation model.
Excessive randomness adversely affects pre-trained models because refinement is not necessary for every pixel in the seismogram. Only higher frequency components require refinement in the Hilbert encoding; for lower frequency pixels, one-grid Hilbert encoding is sufficient.
We aim to develop a more effective strategy for two-grid refinement rather than using the random sampling approach.

\section{Choices of spectrum splitting threshold}
\label{sec:spec}


The high-frequency components are associated with rapid changes in ground motion and correspond to shorter wavelengths. In contrast, low-frequency components relate to slower variations in ground motion and are linked to longer wavelengths.~\cite{yilmaz2001,sheriff1995}
High-frequency components are valuable for fine-structure identification, high-resolution imaging, and lithological analysis. Meanwhile, low-frequency components are useful for deep structure detection, velocity model construction, and full waveform inversion.~\cite{virieux2009,pratt1999}
By integrating high-frequency and low-frequency information, we can enhance multiscale imaging and improve the accuracy of our interpretations.~\cite{fichtner2010,tarantola2005}

\begin{figure*}[htbp]
    \centering
    \includegraphics[width=0.8\textwidth]{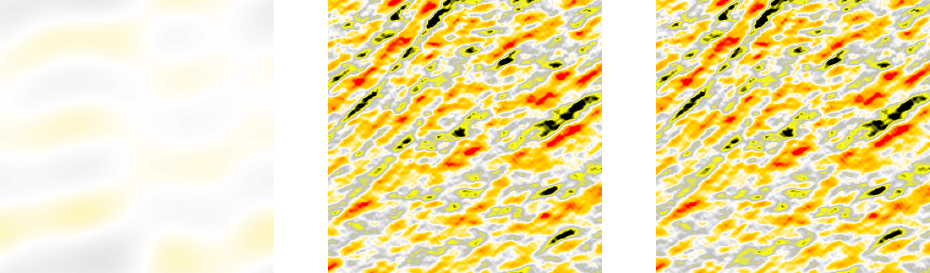}
    \includegraphics[width=0.8\textwidth]{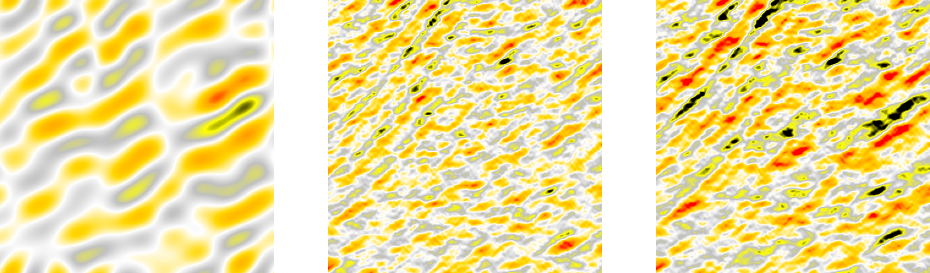}
     \includegraphics[width=0.8\textwidth]{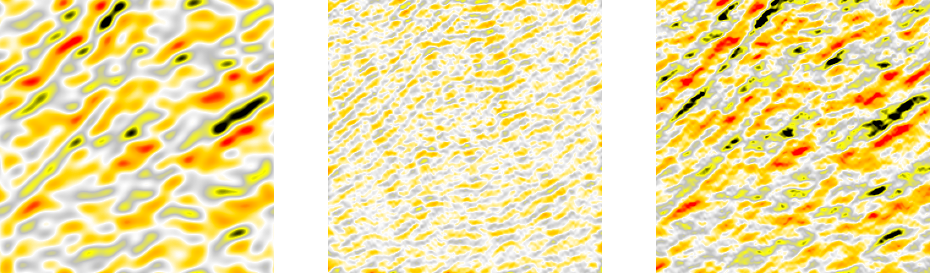}
    \includegraphics[width=0.8\textwidth]{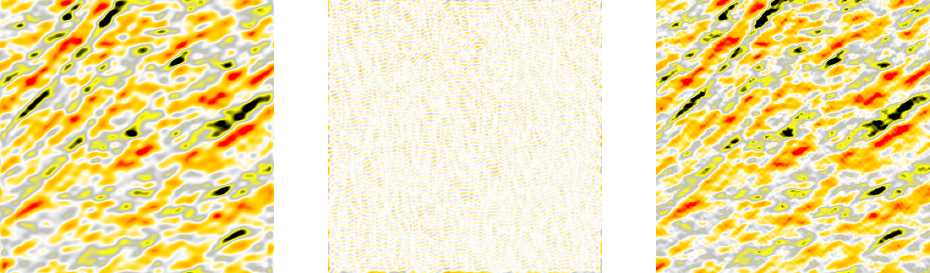}   
     \includegraphics[width=0.8\textwidth]{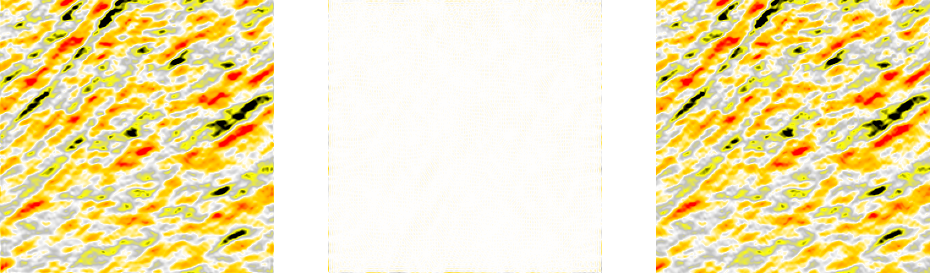}
    \caption{Frequency Decomposition with Different Threshold Example 1.}
    \label{fig:freq_decomp_1}
\end{figure*}

\begin{figure*}[htbp]
    \centering
    \includegraphics[width=0.8\textwidth]{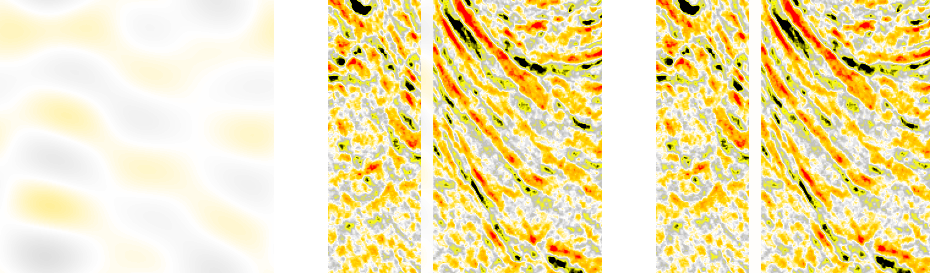}
    \includegraphics[width=0.8\textwidth]{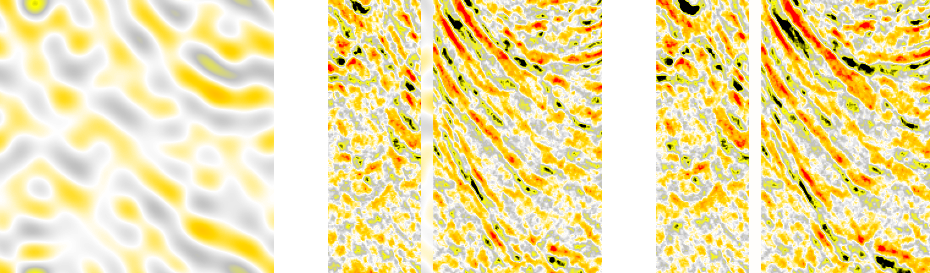}
     \includegraphics[width=0.8\textwidth]{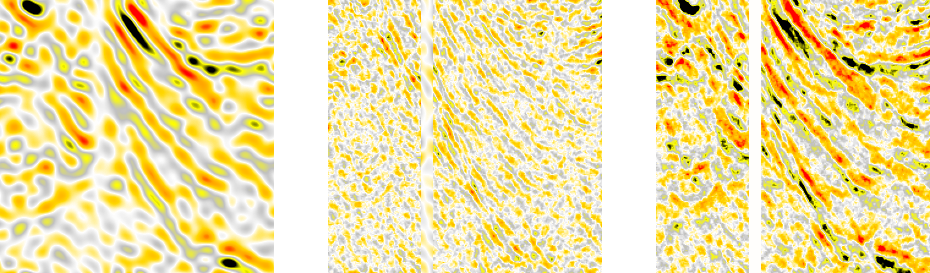}
    \includegraphics[width=0.8\textwidth]{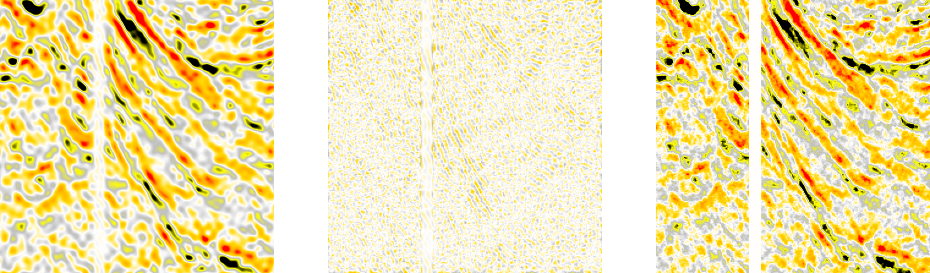}   
     \includegraphics[width=0.8\textwidth]{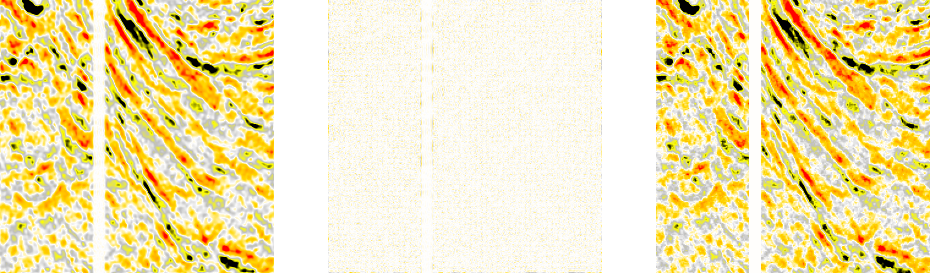}
    \caption{Frequency Decomposition with Different Threshold Example 2.}
    \label{fig:freq_decomp_2}
\end{figure*}

\begin{figure*}[htbp]
    \centering
    \includegraphics[width=0.8\textwidth]{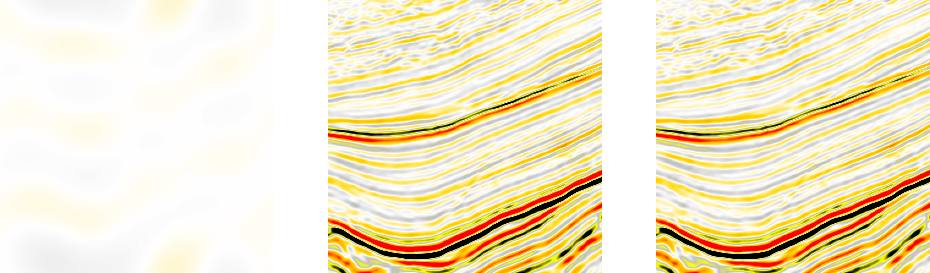}
    \includegraphics[width=0.8\textwidth]{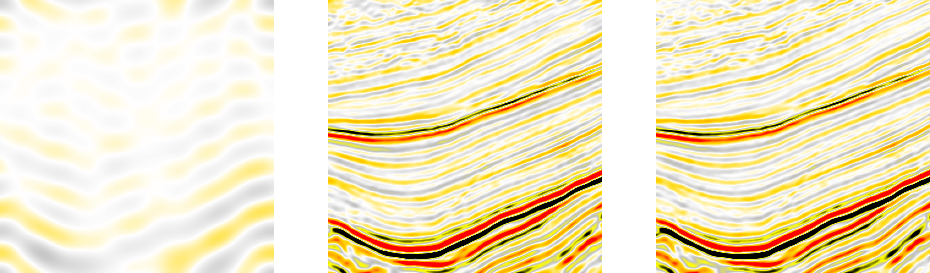}
     \includegraphics[width=0.8\textwidth]{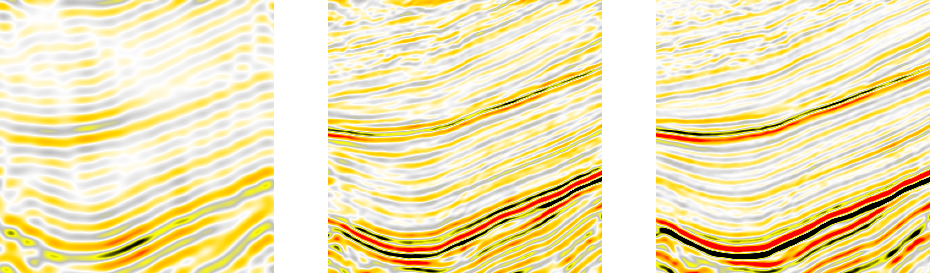}
    \includegraphics[width=0.8\textwidth]{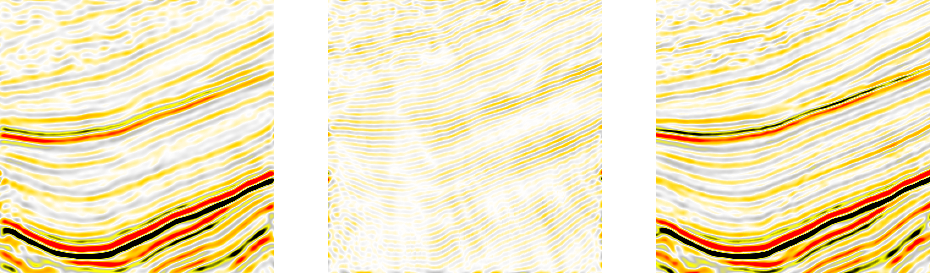}   
     \includegraphics[width=0.8\textwidth]{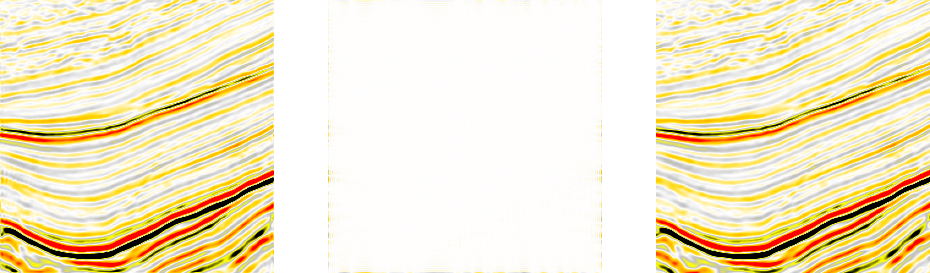}
    \caption{Frequency Decomposition with Different Threshold Example 3.}
    \label{fig:freq_decomp_3}
\end{figure*}

We perform frequency decomposition using different thresholds for high and low frequencies in three seismogram images, as illustrated in Figures~\ref{fig:freq_decomp_1}, \ref{fig:freq_decomp_2}, and \ref{fig:freq_decomp_3}. From left to right, each figure presents the low-frequency, high-frequency, and original images.
From top to bottom, the split threshold is $k_0 \in \{4, 8, 16, 32, 64 \}.$
The first row in Figure~\ref{fig:freq_decomp_1} shows the split results with the threshold $4$. 
The image on the left represents the low-frequency components with frequency $k \leq 4$. 
The medium image represents the high frequency component $k > 4$. 
Moreover, the right one represents the original image by summing up the low- and high-frequency components in the physical domain. 

Figure~\ref{fig:freq_decomp_1} that increasing the frequency split threshold shifts more energy from high- to low-frequency components. However, since both components are crucial for downstream tasks, an even energy distribution between them is desirable. Consequently, we selected a threshold of $k_0 = 16$
 =16 (row 3 in Figure~\ref{fig:freq_decomp_1}) during training of the foundation model, ensuring balanced and informative high and low frequency representations on the seismogram. Additionally, Sect. 4.5 (Table 4) reports the $\ell_2$ norms for both components, further validating this observation.


Furthermore, we record the $\ell_2$ norm for high- and low-frequency compounds in the main context Sect.4, Table 2, which confirms the observation. 
When the threshold is small, for example, $k_0 = 4$, the high frequency occupies the main energy. However, for a large threshold $k_0 = 64$, most energy is concentrated in the low frequency component. 
For $k_0 = 16$, the energy for low frequency is $165.75$ while for high frequency is $139.03$. The high- and low-frequency components show similar energy in this situation. This the main reason why we choose $16$ as the spetrum split threshold.


Similar observations can be observed and analyzed in Figure~\ref{fig:freq_decomp_2} and Figure~\ref{fig:freq_decomp_3}.

\begin{figure*}
    \centering 
     \begin{subfigure}[b]{0.25\textwidth}
         \centering
         \includegraphics[width=\textwidth]{pic/epochs/700_low_1100380.dat.png}
    \end{subfigure}
     \hspace{0.55cm}
     \begin{subfigure}[b]{0.25\textwidth}
         \centering
         \includegraphics[width=\textwidth]{pic/epochs/700_high_1100380.dat.png}
     \end{subfigure}
     \hspace{0.55cm}
     \begin{subfigure}[b]{0.25\textwidth}
         \centering
         \includegraphics[width=\textwidth]{pic/epochs/700_orig_1100380.dat.png}
     \end{subfigure}\\
     \begin{subfigure}[b]{0.25\textwidth}
         \centering
         \includegraphics[width=\textwidth]{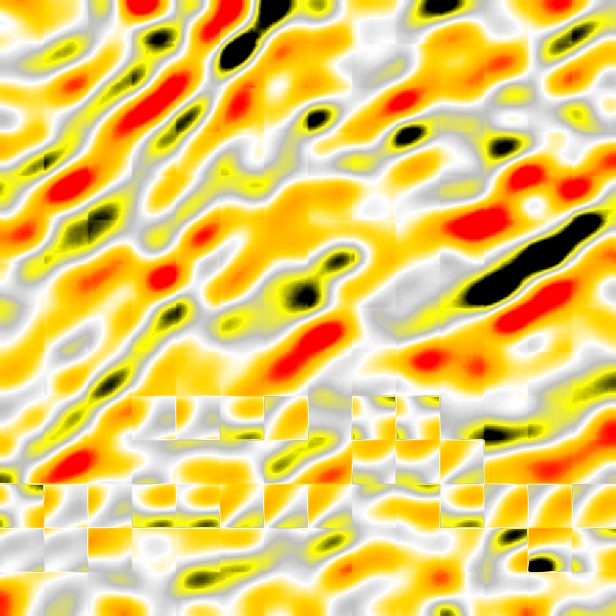}
     \end{subfigure}
     \hspace{0.55cm}
     \begin{subfigure}[b]{0.25\textwidth}
         \centering
         \includegraphics[width=\textwidth]{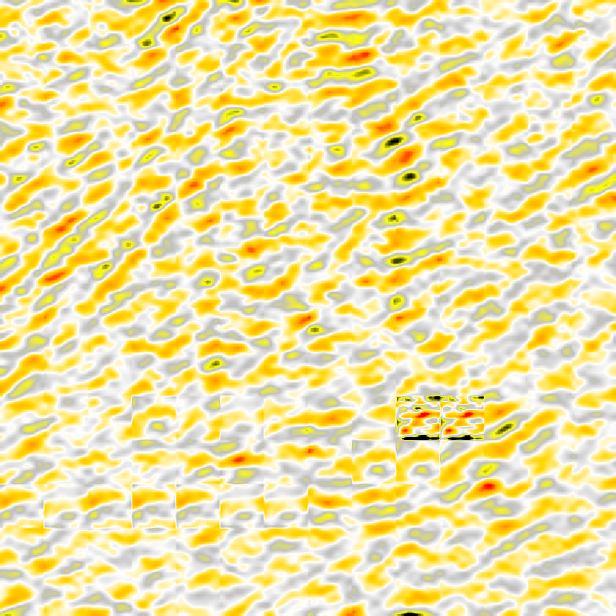}
     \end{subfigure}
     \hspace{0.55cm}
     \begin{subfigure}[b]{0.25\textwidth}
         \centering
         \includegraphics[width=\textwidth]{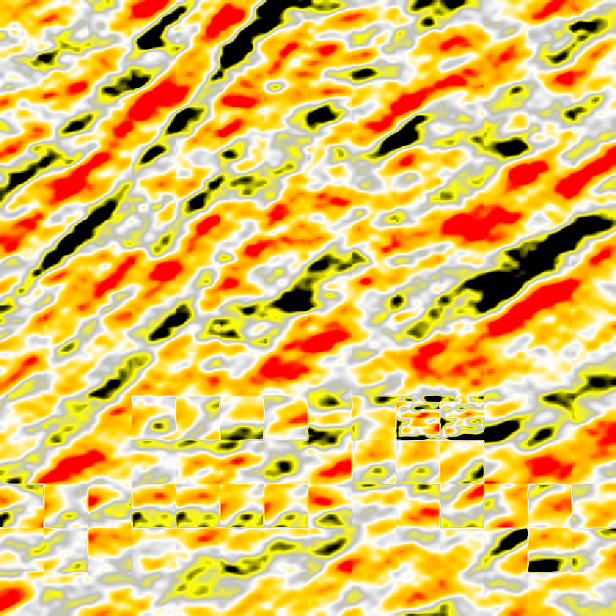}
    \end{subfigure}\\
     \begin{subfigure}[b]{0.25\textwidth}
         \centering
         \includegraphics[width=\textwidth]{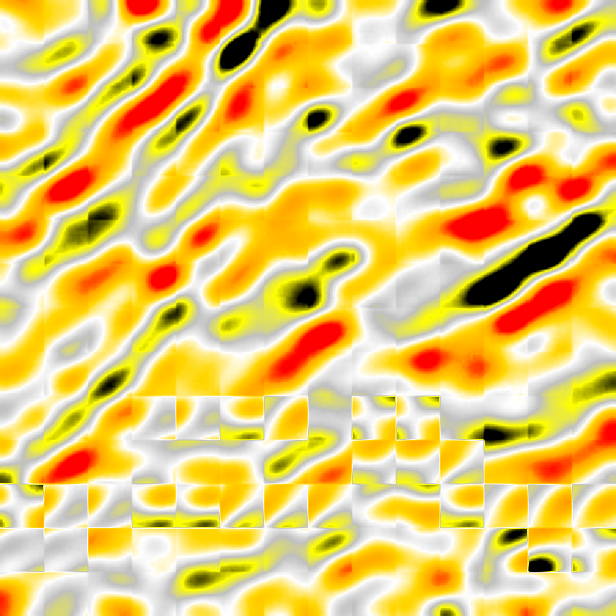}
     \end{subfigure}
     \hspace{0.55cm}
     \begin{subfigure}[b]{0.25\textwidth}
         \centering
         \includegraphics[width=\textwidth]{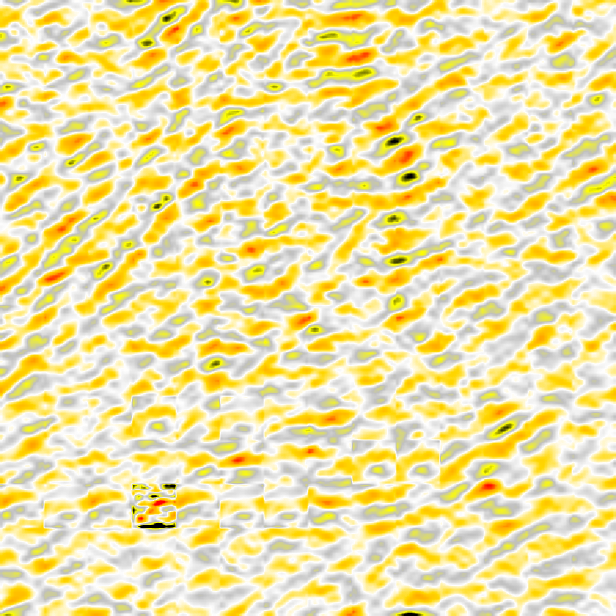}
     \end{subfigure}
     \hspace{0.55cm}
     \begin{subfigure}[b]{0.25\textwidth}
         \centering
         \includegraphics[width=\textwidth]{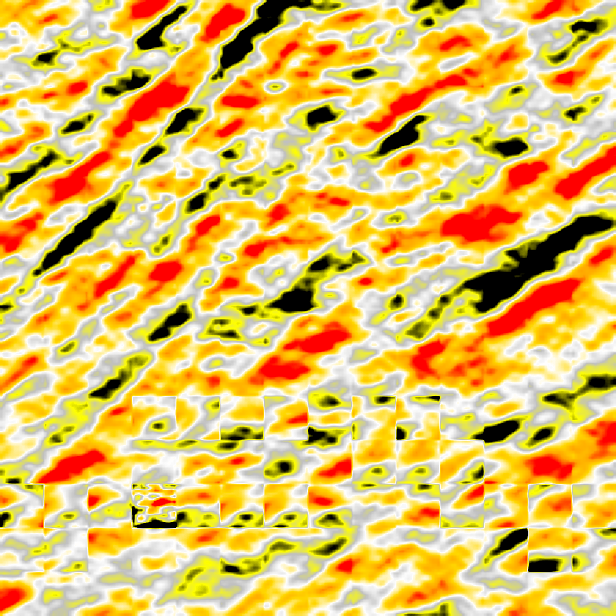}
    \end{subfigure}\\
    \begin{subfigure}[b]{0.25\textwidth}
         \centering
         \includegraphics[width=\textwidth]{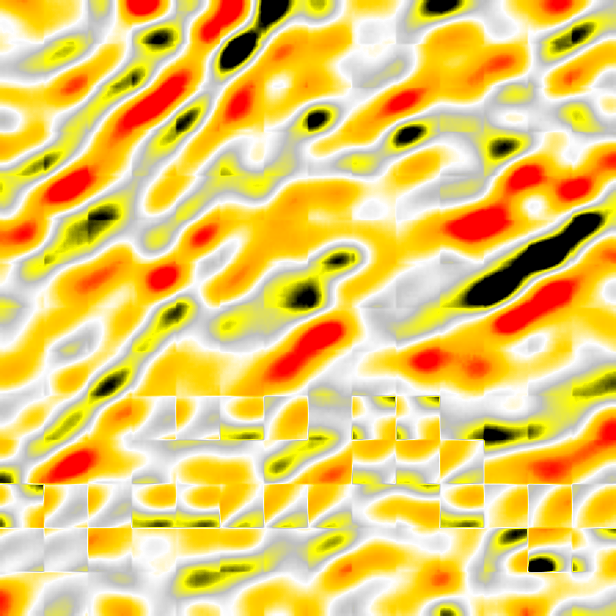}
     \end{subfigure}
     \hspace{0.55cm}
     \begin{subfigure}[b]{0.25\textwidth}
         \centering
         \includegraphics[width=\textwidth]{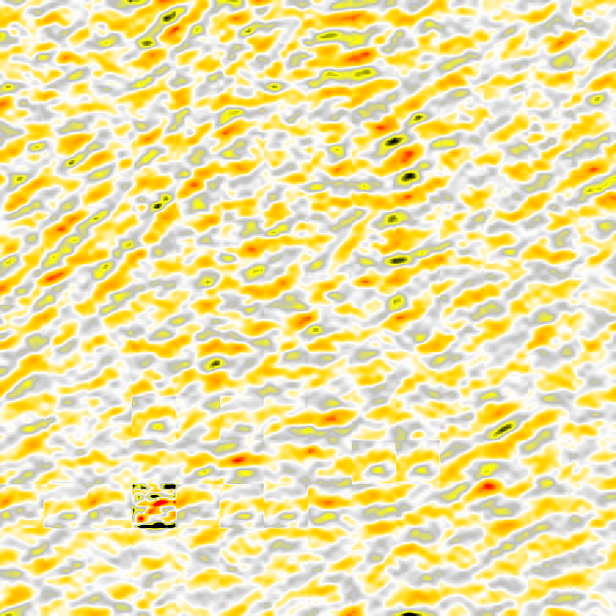}
     \end{subfigure}
     \hspace{0.55cm}
     \begin{subfigure}[b]{0.25\textwidth}
         \centering
         \includegraphics[width=\textwidth]{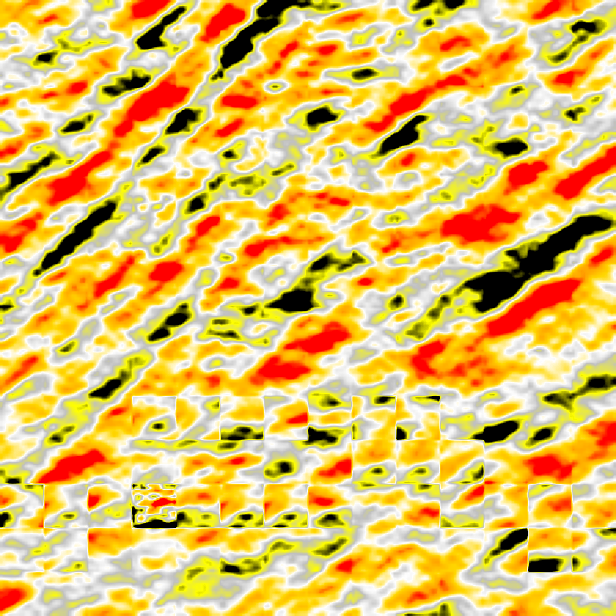}
    \end{subfigure}\\
    \begin{subfigure}[b]{0.25\textwidth}
         \centering
         \includegraphics[width=\textwidth]{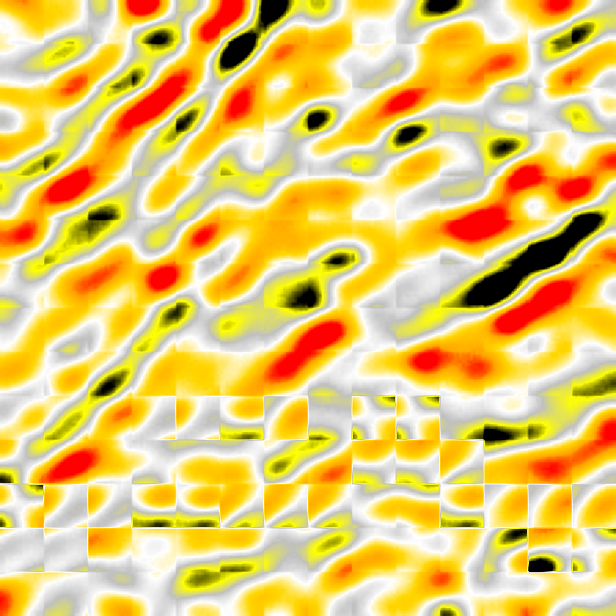}
     \end{subfigure}
     \hspace{0.55cm}
     \begin{subfigure}[b]{0.25\textwidth}
         \centering
         \includegraphics[width=\textwidth]{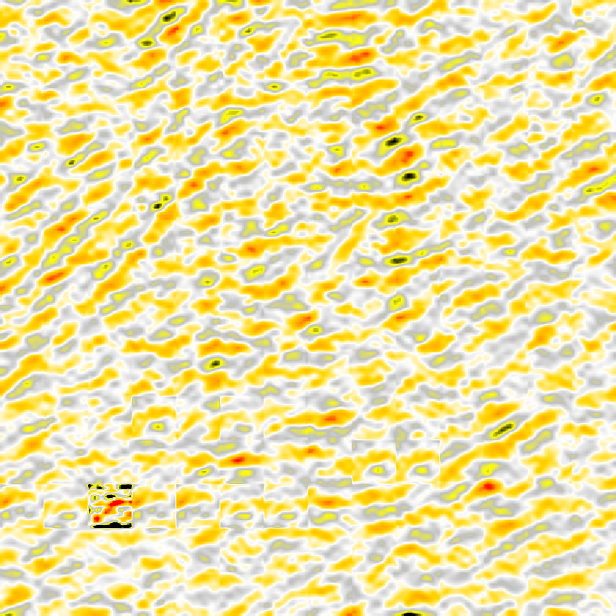}
     \end{subfigure}
     \hspace{0.55cm}
     \begin{subfigure}[b]{0.25\textwidth}
         \centering
         \includegraphics[width=\textwidth]{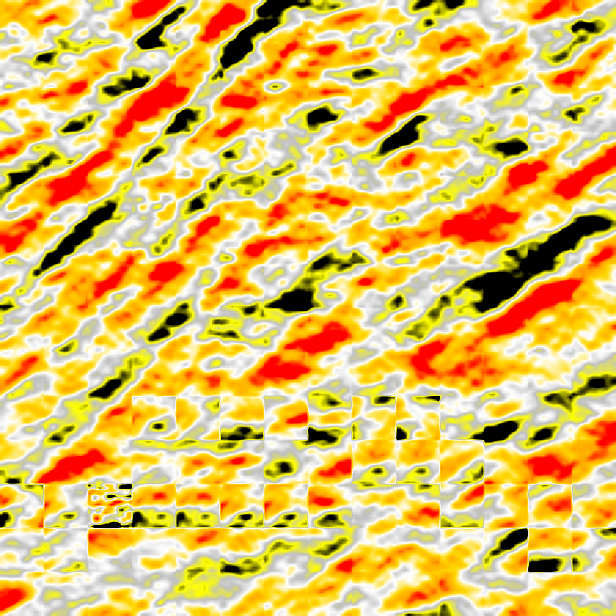}
    \end{subfigure}\\
\caption{Reconstruction with Different Timesteps. The left column represents the low-frequency components, the middle column represents the high-frequency components, and the right column displays the reconstructed components, which are a combination of both high- and low-frequency components. From the top to bottom, we show the reconstruction results with epochs $700, 1000, 1100, 1300$ and $1500$ respectively.}
\label{fig:timestep}
\end{figure*}

\section{Reconstruction with Different Timesteps}
\label{sec:recons_time}

In this section, we visualize the reconstruction results in different time steps, as shown in Figure~\ref{fig:timestep}. The left column represents the low-frequency components, the middle column represents the high-frequency components, and the right column displays the reconstructed components, which are a combination of high- and low-frequency elements. The results are organized from top to bottom, corresponding to training epochs of 700, 1000, 1100, 1300, and 1500.

From Figure~\ref{fig:timestep}, we can see that by epoch 700, we already achieve a satisfactory reconstruction of the high- and low-frequency components. As training epochs increase, more details are captured in both frequency domains, indicating that additional training improves the quality of detail reconstruction. However, by epoch 1300, we notice some unwarranted details in the high-frequency reconstruction. This may be a result of overfitting, where patterns learned from the low-frequency components mislead the reconstruction of the high-frequency components.

\section{Downstream Tasks}

We further compare our methods (\modelname) with \base~~\cite{sheng2025seismic} on four downstream tasks. The comparison are shown in the following table (Table~\ref{tab:downstream}). 
The following table shows that \modelname~improves slightly compared to \base~in the downstream tasks including Denoise and Interpolation. 
\begin{table}[ht]
\centering
\caption{Performance comparison of \modelname~and \base~\cite{sheng2025seismic} on denoising and interpolation tasks.}
\label{tab:downstream}
\begin{tabular}{lcccccc}
\toprule
Model & \multicolumn{3}{c}{Denoise} & \multicolumn{3}{c}{Interpolation} \\
\cmidrule(r){2-4} \cmidrule(l){5-7}
      & Task & MSE   & PSNR  & Task & MSE   & PSNR   \\
\midrule
\modelname & denoise & 0.758 & 7.066 & inter & 0.173 & 18.578 \\
\base~\cite{sheng2025seismic}  & denoise & 0.776 & 6.922 & inter & 0.218 & 17.411 \\
\bottomrule
\end{tabular}
\end{table}

\end{document}